\documentclass[twocolumn]{svjour3}          
\smartqed  
\usepackage{graphicx}
\usepackage{epstopdf}
\usepackage{hyperref}
\usepackage{color}
\usepackage{soul}
\graphicspath{ {./figures4paper1/}}
\usepackage{graphicx}
\usepackage{caption}
\usepackage{subcaption}
\usepackage{amsmath}
\usepackage{amssymb}
\usepackage{algorithm,algorithmic}
\usepackage{paralist}
\usepackage{multirow}
\DeclareMathOperator*{\argmax}{arg\,max}
\usepackage[numbers,sort&compress]{natbib}

\journalname{J Math Imaging Vis}
\begin{document}

\title{Analysis of Vessel Connectivities in Retinal Images \\ by Cortically Inspired Spectral Clustering
}

\titlerunning{Cortically Inspired Connectivity Analysis in Retinal Images}        

\author{Marta Favali \and
        Samaneh Abbasi-Sureshjani \thanks{M. Favali and S. Abbasi-Sureshjani contributed equally to this work.}
	 \and \\  Bart ter Haar Romeny \and Alessandro Sarti 
}


\institute{M. Favali \and A.Sarti 
		\at Centre d'Analyse et de Math\'ematique Sociales, Ecole des Hautes \'Etudes en Sciences Sociales, 190-198, Avenue de France 75244 Paris Cedex 13  France\\
              \email{\{marta.favali, alessandro.sarti\}@ehess.fr}           
           \and 
           S. Abbasi-Sureshjani \at
              Department of Biomedical Engineering, Eindhoven University of Technology, P.O. Box 513, 5600 MB Eindhoven, The Netherlands \\
              \email{s.abbasi@tue.nl}         
               \and
              B. ter Haar Romeny  \at
			Department of Biomedical and Information Engineering,
			Northeastern University, 500 Zhihui Street, Shenyang,
			110167 China\\
			\email{bart.romeny@outlook.com}   		  
}

\date{Received: 23 October 2015 / Accepted:  08 February 2016}

\maketitle

\begin{abstract}
Retinal images provide early signs of diabetic retinopathy, glaucoma and hypertension. These signs can be investigated based on microaneurysms or smaller vessels. The diagnostic biomarkers are the change of vessel widths and angles especially at junctions, which are investigated using the vessel segmentation or tracking. Vessel paths may also be interrupted; crossings and bifurcations may be disconnected. This paper addresses a novel contextual method based on the geometry of the primary visual cortex (V1) to study these difficulties. We have analysed the specific problems at junctions with a connectivity kernel obtained as the fundamental solution of the Fokker-Planck equation, which is usually used to represent the geometrical structure of multi-orientation cortical connectivity. By using the spectral clustering on a large local affinity matrix constructed by both the connectivity kernel and the feature of intensity, the vessels are identified successfully in a hierarchical topology each representing an individual perceptual unit.
\end{abstract}

\section{Introduction}
\label{intro}

\sloppy \paragraph{Clinical importance of retinal blood vessels:} 
Epidemic growth of systemic, cardiovascular and ophthalmologic diseases such as diabetes, hypertension, glaucoma, and arteriosclerosis ~\cite{mozaffarian2015heart,tabish2007diabetes,hu2011globalization}, their high impact on the quality of life, and the substantial need for increase in health care resources~\cite{viswanath2003diabetic,ICOGuidelines} indicate the importance of conducting large screening programs for early diagnosis and treatment of such diseases. This is impossible without using automated computer-aided systems because of the large population involved.

The retina is the only location where blood vessels can be directly monitored non-invasively in vivo~\cite{xu2010novel}. 
Retinal vessels are connected and form a treelike structure. The local orientation and intensity of vessels change gradually along their lengths; however, these local properties may vary highly for different vessels.
Studies show that the quantitative measurement of morphological and geometrical attributes of retinal vasculature, such as vessel diameter, tortuosity, arteriovenous ratio and branching pattern and angles are very informative in early diagnosis and prognosis of several diseases~\cite{lowell2004measurement,foracchia2001extraction,chapman2002peripheral,smith2004retinal,hubbard1999methods,habib2014association,wasan1995vascular}. More specifically, since two blood vessels with the same type never cross each other and arteries are more likely to cross over veins, studying the behaviour of blood vessels at crossings for detecting the branch retinal vein occlusion (occlusion of the vein) and arteriovenous nicking helps in diagnosis of hypertension (increased arterial blood pressure), arteriosclerosis (hardening of arteries) and stroke~\cite{wong2007eye,wong2001retinal,fruttiger2007development}.
The other clinically highly promising but still underrated information is based on studying the smaller vessels, because it is expected that the signs of diseases such as diabetic retinopathy appears in smaller vessels earlier than in larger ones. 

\sloppy\paragraph{Vessel extraction and its difficulties:}The vasculature can be extracted by means of either pixel classification or vessel tracking. Several segmentation and tracking methods have been proposed in the literature~\cite{Fraz2012407,buhler2004geometric,felkel2001vessel}. 
In pixel classification approaches image pixels are labeled either as vessel or non-vessel pixels. Therefore, a vessel likelihood (soft segmentation) or binary map (hard segmentation) is created for the retinal image. Although the vessel locations are estimated in these approaches, they do not provide any information about vessel connectivities. On the contrary, in tracking based approaches, several seed points are selected and the best connecting paths between them are found~\cite{al2009active,chutatape1998retinal,poon2007live,quek2001vessel,xu2011vessel,can1999rapid,de2014tracing,de2013automated,delibasis2010automatic,gonzalez2010delineating}. The main benefit of vessel tracking approaches is that they work at the level of a single vessel rather than a single pixel and they try to find the best path that matches the vessel profile. Therefore, the information extracted from each vessel segment (e.g. diameter and tortuosity) is more accurate and reliable. 

There are several difficulties for both vessel segmentation and tracking approaches.
 Depending on imaging technology and conditions, these images could be affected by noise in several degrees. Moreover, non-uniform luminosity, drift in image intensity, low contrast regions and also central vessel reflex make the vessel detection and tracking complicated. Several image enhancement, normalization and denoising techniques have been developed to tackle these complications (e.g.~\cite{abbasi2015biologically,Foracchia2005,narasimha2008improved}). 
 
\begin{figure*}
  \centering 
  \includegraphics[width=4 in]{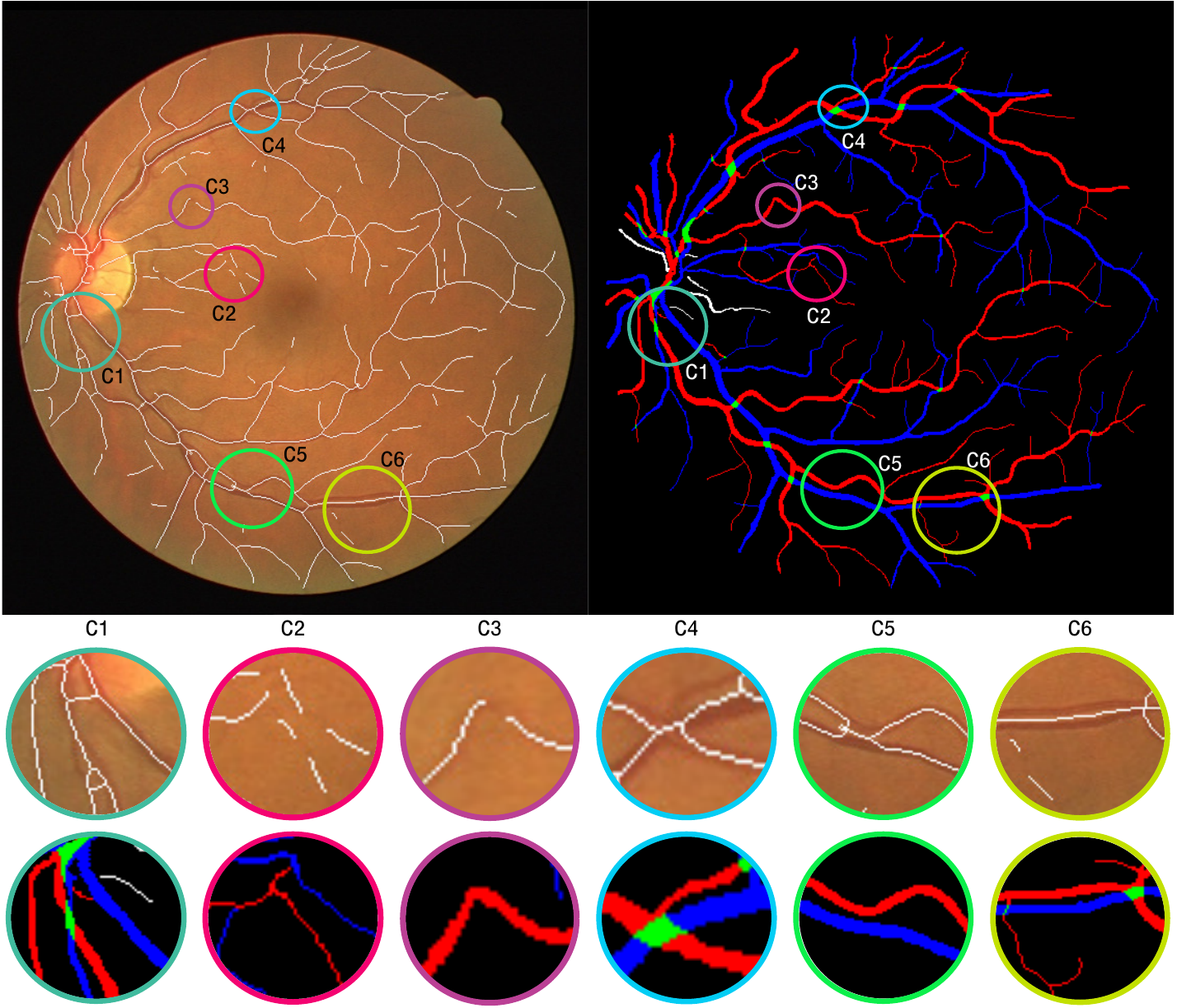} 
  \caption{A sample image selected from the DRIVE dataset~\cite{staal2004} with centrelines extracted by applying the morphological skeletonization on the segmented image (top left), and its corresponding artery/vein ground truth from the RITE dataset~\cite{hu2013automated} (top right). Several difficult cases are highlighted in this figure. C1: complex junction with the presence of a bifurcation and a crossing with a narrow crossing angle; C2: interrupted lines and missing small vessels; C3: high curvature vessel; C4: complex junction, a crossing and bifurcation, wrong thinning at crossing; C5: two nearby parallel vessels merged as one group; C6: missing small vessel, merged parallel vessels and  interrupted segment.}
  \label{fig:trackingIssues}
\end{figure*}

The tracking methods are often performed exploiting the skeleton of the segmented images. Thus, non-perfect segmentation or wrong skeleton extraction results in topological tracing errors e.g. disconnections and non-complete subtrees as discussed in several methods proposed in the literature~\cite{joshi2011automated,al2009active,de2013automated,de2014tracing,gonzalez2010delineating}. Typical non-perfections include missing small vessels, wrongly merged parallel vessels, disconnected or broken up vessel segments and the presence of spur branches in thinning. 
Moreover, the greater difficulty arises at junctions and crossovers: small arteriovenous crossing angles, complex junctions when several junctions are close together, or presence of a bifurcation next to a crossing makes the centerline extraction and tracing challenging. These difficulties are mentioned as the tracking limitations in the literature. Some of these challenging cases are depicted in Fig.~\ref{fig:trackingIssues} with their corresponding artery/vein ground truth labels. Arteries and veins are annotated in red and blue colors respectively. The green color represents the crossing and the types of the white vessels are not known.  

\paragraph{ Gestalt theory and cortically-inspired spectral clustering:}

\begin{figure}[htbp]
	\centering 
	\begin{subfigure}[b]{1 in} \centering
		\includegraphics[trim= 0.3in 0.5in 0.45in 0.2in, clip, width=1 in]{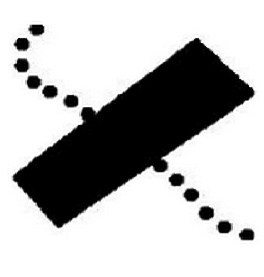}
		\caption{}
		\label{fig:goodCont}
	\end{subfigure}
		\begin{subfigure}[b]{1 in} \centering
			\includegraphics[trim= 0.3in 0.3in 0.3in 0.2in, clip, width=1 in]{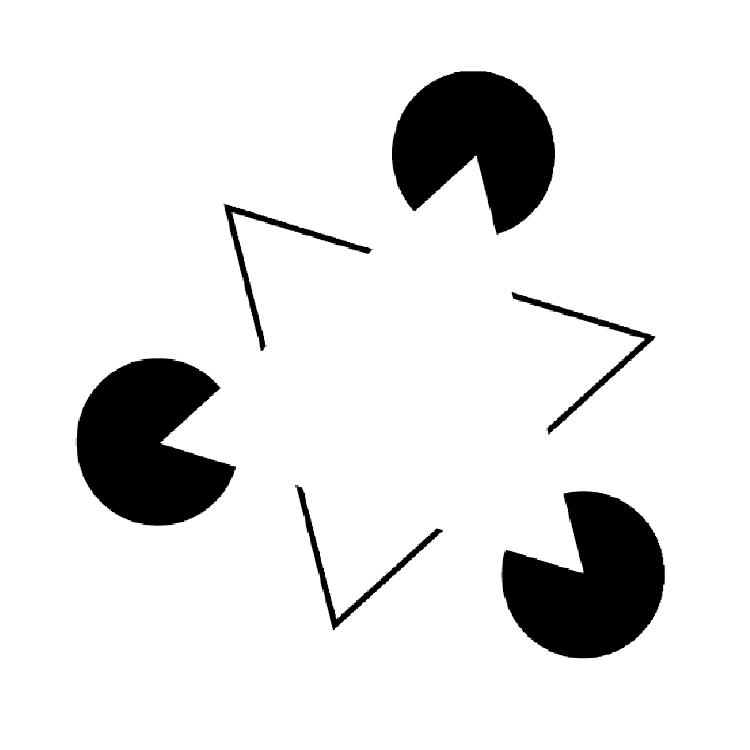}
				\caption{}
				\label{fig:closure}
		\end{subfigure}
		\begin{subfigure}[b]{1 in} \centering
			\includegraphics[trim= 1.5in 3in 1.5in 1.5in, clip, width=1 in]{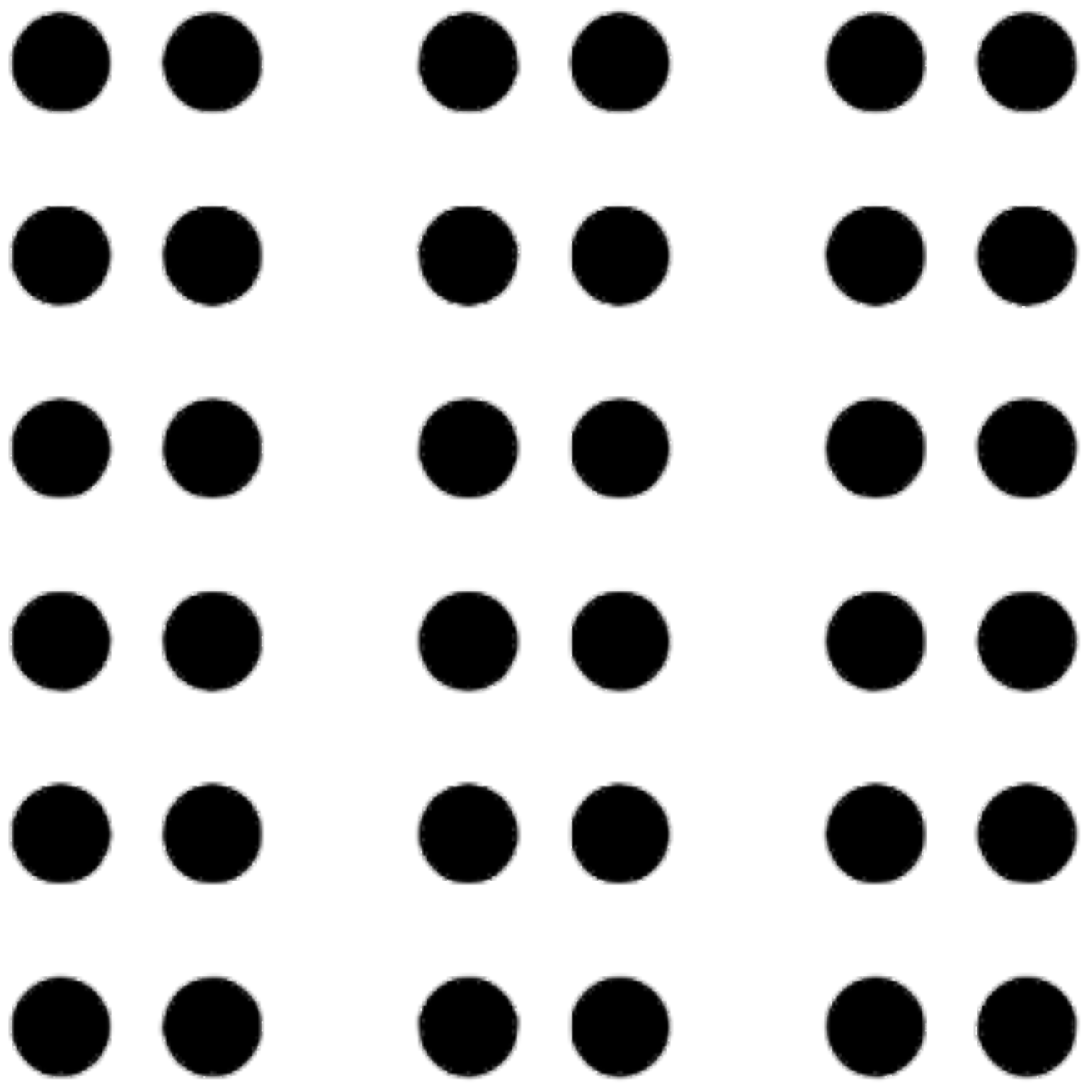}
				\caption{}
				\label{fig:proximity}
		\end{subfigure}
	\caption{Examples of (\subref{fig:goodCont}) good continuation, (\subref{fig:closure}) closure and (\subref{fig:proximity}) proximity Gestalt laws.}\label{fig:good}
\end{figure}

Visual tasks like image segmentation and grouping can be explained with the theory of the Berliner Gestalt psychology, that proposed local and global laws to describe the properties of a visual stimulus \cite{wagemans2012century,wertheimer1938laws}. In particular, the laws of good continuation, closure and proximity have a central role in the individuation of perceptual units in the visual space, see Fig.~\ref{fig:good}.
In \cite{parent1989trace} perceptual grouping was considered to study the problem of finding curves in biomedical images.
In order to study the property of good continuation, Field, Hayes and Hess introduced in \cite{field1993contour} the concept of an association field, that defines which properties the elements of the stimuli should have to be associated to the same perceptual unit, such as co-linearity and co-circularity. 
In \cite{bosking1997orientation} Bosking showed how the rules of association fields are implemented in the primary visual cortex (V1), where neurons with similar orientation are connected with long-range horizontal connectivity. A geometric model of the association fields based on the functional organization of V1 has been proposed in \cite{citti2006cortical}. This geometric approach is part of the research line proposed by \cite{koenderink1987representation,hoffman1989visual,mumford1994elastica,zucker2006differential,sarti2008symplectic,petitot1999vers} and applications to image processing can be found in \cite{boscain2012anthropomorphic,duits2010left,duits2013evolution}.  

In this work, a novel mathematical model based on this geometry has been applied to the analysis of retinal images to overcome the above mentioned connectivity problems in vessel tracking. The proposed method represents an engineering application of segmenting  and representing blood vessels inspired by the modeling of the visual cortex. This shows how these models can be applied to the analysis of medical images and how these two fields can be reciprocally used to better understand and reinforce each other. 

This method, which is not dependent on centreline extraction, is based on the fact that in arteriovenous crossings there is a continuity in orientation and intensity of the artery and vein respectively, i.e. the local variation of orientation and intensity of individual vessels is very low. 
The proposed method models the connectivity as the fundamental solution of the Fokker-Planck equation, which matches the statistical distribution of edge co-occurrence in natural images and is a good model of the cortical connectivity~\cite{sanguinetti2008image}. 

Starting from this connectivity kernel and considering the Euclidean distance between intensities of blood vessels, we build the normalized affinity matrix. Since at crossings the vessels have different intensities (types), including this feature in the construction of the affinity matrix adds more discriminative information. This spectral approach, first proposed for image processing, is inspired by \cite{perona1998factorization,coifman2006diffusion,shi2000normalized,weiss1999segmentation}. Moreover, recent results of \cite{sarti2015constitution} show how the spectral analysis could actually be implemented by the neural population dynamics of the primary visual cortex (V1).
Finally, we use a spectral clustering algorithm to find and group the eigenvectors linked to the highest eigenvalues of the affinity matrix. We will describe how these groups represent different perceptual units (vessels) in retinal images. Originally, the spectral clustering was exploited for good continuation, closure and proximity, see Fig.~\ref{fig:good}. It excelled in finding connections, e.g. broken vessel segments. In this paper, we go further by solving many challenges at vessel crossings and bifurcations. 

\paragraph{Paper structure:} The remainder of the article is organized as follows. In Sect. \ref{sec:geometryV1}, after describing the geometry of the functional architecture of the primary visual cortex, it is explained in detail how to lift the stimulus in cortical space using a specific wavelet transform. This lifting converts the image from the space of positions ($R^2$) to the joint space of positions and orientations ($R^2\times S^1$). Then the connectivity kernel and the construction of the affinity matrix based on this kernel and the intensity is described. Next step is the spectral clustering  algorithm, that is used to extract the grouping information of the stimuli, explained in Sect. \ref{sec:specClust}. The experiments applied on retinal images are explained step by step and the results are presented in Sect. \ref{sec:experiments}.
Finally, the paper is concluded in Sect. \ref{sec:conclusion} by briefly summarizing the proposed method, discussing its strengths in preserving the connectivities in the retinal vasculature network and proposing potential improvements as future work. 

\section{Geometry of Primary Visual Cortex}
\label{sec:geometryV1}

%
%
\subsection{Lifting of the Stimulus in the Cortical Space}
\label{sec:os}

In this section we recall the structure of the geometry of the primary visual cortex (V1). Hubel and Wiesel \cite{hubel1977ferrier} first discovered that the visual cortex is organized in a hypercolumnar structure, where each point corresponds to a simple cell sensitive to a stimulus positioned in (x,y) and orientation $\theta$. 
In other words, simple cells  extract the orientation information at all locations and send a multi-orientation field to higher levels in the brain. Also, it is well known that objects with different orientations can be identified by the brain even when they are partly occluded, noisy or interrupted~\cite{hubel1962receptive}.

Motivated by these findings, a new transformation was proposed \cite{duits2007invertible,duits2007image},to lift all elongated structures in 2D images to a new space of positions and orientations ($R^2\times S^1$) using elongated and oriented wavelets. By lifting the stimulus, multiple orientations per position could be detected. Thus, crossing and bifurcating lines are disentangled into separate layers corresponding to their orientations.

\begin{figure*}[htbp]
\centering
	\begin{subfigure}{1 in}
	\centering
		\includegraphics[width=1 in]{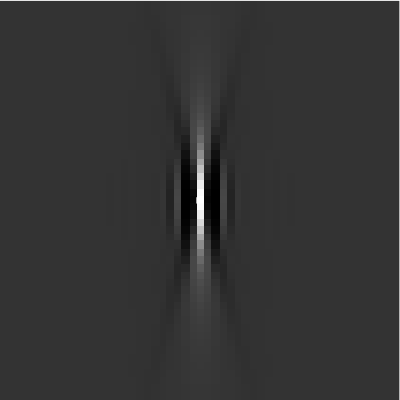} 
		\includegraphics[width=1 in]{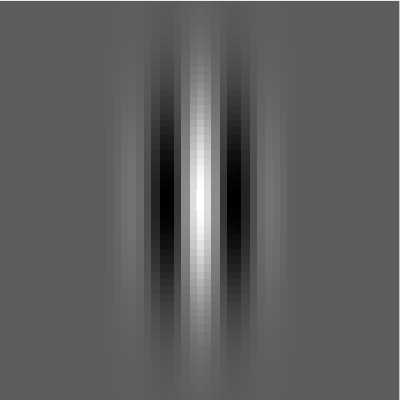} 
		\caption{} \label{fig:waveletReal}
	\end{subfigure}
	\begin{subfigure}{1 in}
	\centering
		\includegraphics[width=1 in]{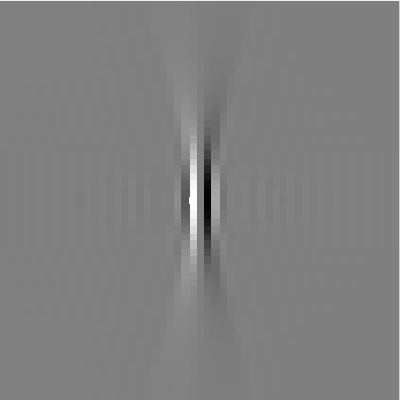} 
		\includegraphics[width=1 in]{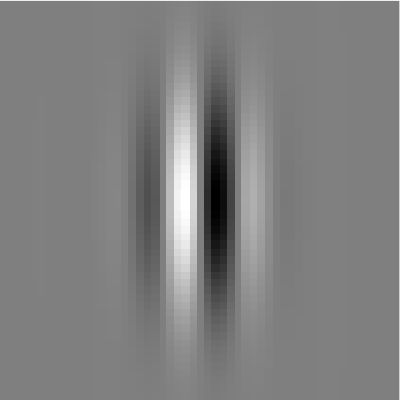} 
		\caption{} \label{fig:waveletImg}
	\end{subfigure}
	\begin{subfigure}{1 in}
	\centering
		\includegraphics[width=1 in]{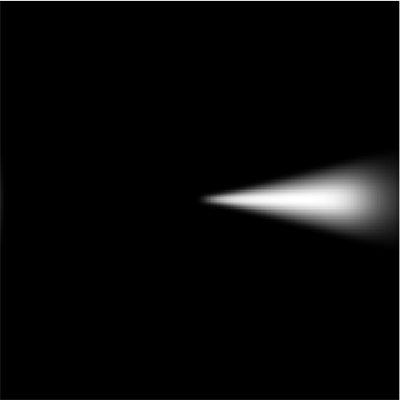} 
		\includegraphics[width=1 in]{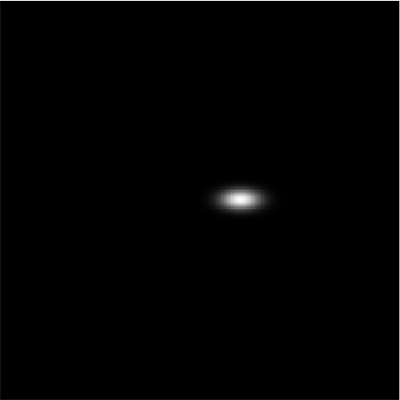} 
		\caption{} \label{fig:waveletFourier}
	\end{subfigure}
	\begin{subfigure}{1 in}
	\centering
		\includegraphics[width=1 in]{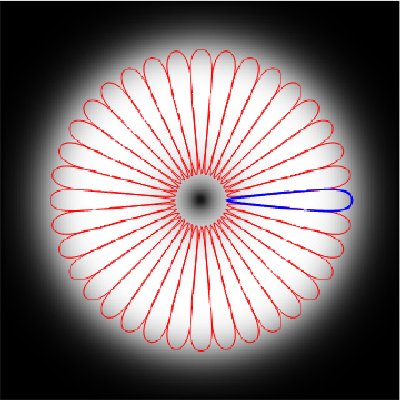} 
		\includegraphics[width=1 in]{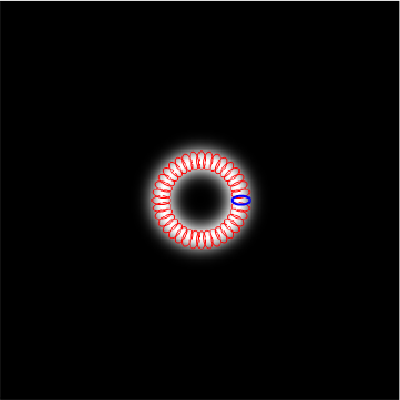} 
		\caption{} \label{fig:waveletOverlay}
	\end{subfigure}
\caption{Visual comparison between cake and Gabor wavelets (top and bottom rows respectively).~(\subref{fig:waveletReal}) and~(\subref{fig:waveletImg}) real and imaginary parts of the wavelets in spatial domain,~(\subref{fig:waveletFourier}) the wavelet in Fourier domain, and~(\subref{fig:waveletOverlay}) the illustration of Fourier space coverage by wavelets in 36 orientations.}
\label{fig:cakeGaborwavelets}
\end{figure*}

Constructing this higher dimension structure simplifies the higher level analysis. However, such representation is constrained and the invertibility of the transformation needs to be guaranteed using the right wavelet. Cake wavelets as a class of proper wavelets~\cite{duits2007invertible,duits2007image,fuhr2005abstract} satisfy this constraint and avoid loss of information. Cake wavelets are directional wavelets similar to Gabor wavelets and have a high response on oriented and elongated structures. Moreover, similar to Gabor wavelets, they have the quadratic property; so the real part contains information about the symmetric structures, e.g. ridges, and the imaginary part contains information about the antisymmetric structures, e.g. edges. Although blood vessels can have several scales, by using the cake wavelets multi-scale analysis is not needed, because they capture the information at all scales. A visual comparison between Gabor and cake wavelets is presented in Fig.~\ref{fig:cakeGaborwavelets}. As seen in this figure, by using the cake wavelet in all orientations, the entire frequency domain is covered; while the Gabor wavelets, depending on their scale cover a limited portion of Fourier space. Therefore, they are scale dependent. The reader is referred to~\cite{Bekkers:2014aa} for more detail.
                                                                                        
\begin{figure*}
	\centering
	\begin{subfigure}[b]{2.8 in}
		\includegraphics[trim= 0.3in 0.6in 0.3in 0.5in, clip, width=2.3 in]{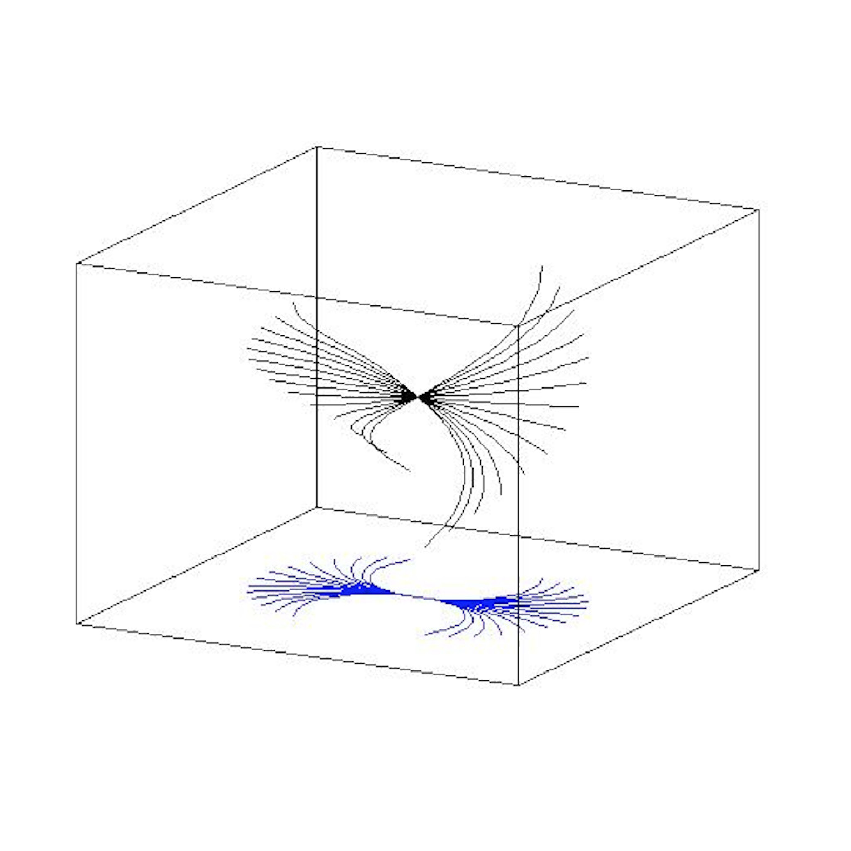}  \caption{Integral curves}
		\label{fig:ker3}
	\end{subfigure}
	\begin{subfigure}[b]{3 in}
		\includegraphics[trim= 0.3in 0.6in 0.3in 0.5in, clip, width=2.5 in]{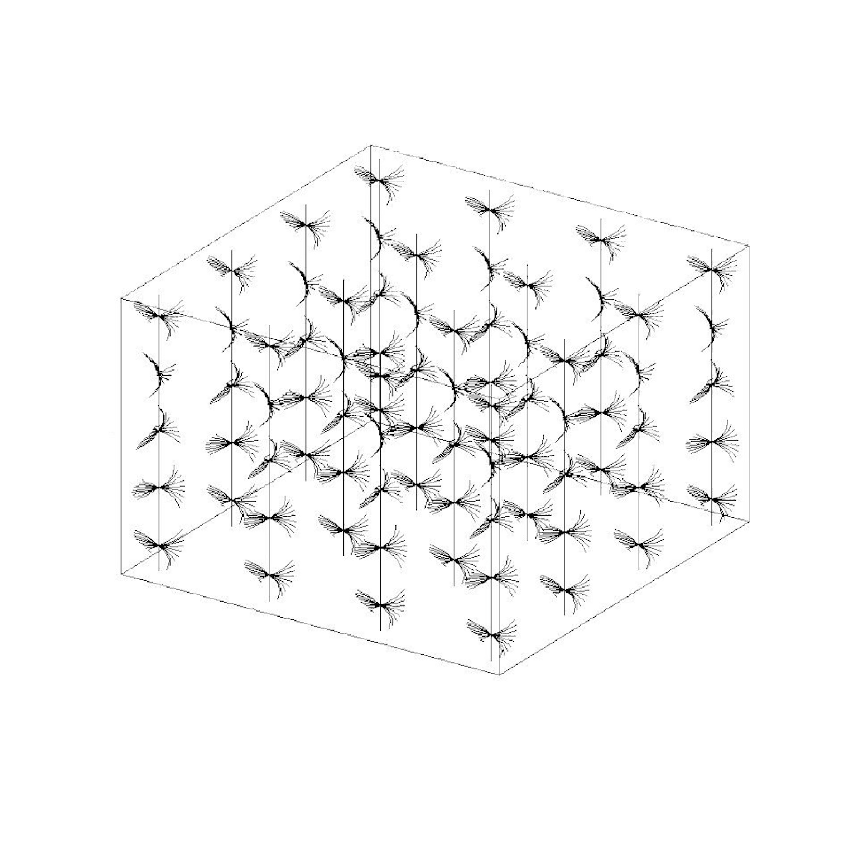} \caption{Distribution of integral curves}
		\label{fig:conn}
	\end{subfigure}

	\caption{
		\subref{fig:ker3}) The integral curves of the vector fields $X_1$ and $X_2$ in the (x,y,$\theta$) space. In blue the projections of the integral curves on the $xy$ plane.
		\subref{fig:conn}) The distribution of the integral curves, modeling the connectivity between points.
		}
	\label{fig:proj}
\end{figure*}

For lifting the stimulus and constructing the higher dimension representation, $U_f: SE(2) \to \mathbb{C}$, the input image $f(x,y)$ is correlated with the anisotropic cake wavelet $\psi$~\cite{duits2007image,duits2007invertible,franken2009crossing}:
 \begin{equation} 
 \begin{split}
U_f(\mathbf{x},\theta)=(\mathcal{W}_\psi [f])(\mathbf{x},\theta) =(\overline{R_\theta(\psi)} \star f)(\mathbf{x}) \\
=\int_{\mathbb{R}^2} \overline{\psi(R^{-1}_\theta (\mathbf{y}-\mathbf{x}))}f(\mathbf{y})\mathrm{d}\mathbf{y}
 \end{split}
 \label{eq:OST}
 \end{equation}
 where $R_\theta$ is the 2D counter-clockwise rotation matrix and the overline denotes the complex conjugate. 
Using this transformation and considering only one orientation per position (the orientation with highest transformation response), the points of a curve $\gamma = (x,y)$ are lifted to new cortical curves and are described in the space $(x,y,\theta)$:
\begin{equation*}
	(x,y) \rightarrow (x,y,\theta).
\end{equation*} 
These curves have been modelled in \cite{citti2006cortical} as integral curves of suitable vector fields: 
\begin{equation}
	\vec X_1 = (\cos\theta,\sin\theta,0),  \mbox{\quad}     \vec X_2 = (0,0,1).
	\label{vector}
\end{equation}
The points of the lifted curves are connected by integral curves of these two vector fields such that:
\begin{equation}
	\gamma : \mathbb{R}\rightarrow SE(2), \mbox{\qquad} \gamma(s)=(x(s),y(s),\theta(s))
	\label{eq1}
\end{equation}
\begin{equation*}
	\gamma'(s)=(k_1(s)\vec X_1 + k_2(s)\vec X_2)(\gamma(s)),\mbox{\quad} \gamma(0)=0. 
\end{equation*}
These curves projected on the 2D cortical plane represent a good model of the association fields, as described in \cite{citti2006cortical} (see Fig.~\ref{fig:proj}).

In order to include the intensity term, we use the Euclidean distance between the intensities of two corresponding points. If $f(x,y)$ represents the image intensity at position $(x,y)$ the stimulus is lifted to the extended 4-dimensional feature space:
\begin{equation*}
	(x,y,\theta) \rightarrow (x,y,\theta,f(x,y)).
\end{equation*}
An admissible curve in this space is defined as the solution of the following differential equation:
\begin{equation}
	\gamma'(s) = (k_1(s)\vec X_1 + k_2(s)\vec X_2 + k_4(s)\vec X_4)(\gamma(s)) 
\end{equation}
\begin{equation*}
	\gamma(0) = (x_1,y_1,\theta_1,f_1) , \mbox{\qquad} \gamma(1) = (x_2,y_2,\theta_2,f_2) 
\end{equation*}
where the vector fields are:
\begin{equation}
	\vec X_1 = (\cos\theta,\sin\theta,0,0),  \mbox{\quad}     \vec X_2 = (0,0,1,0), \mbox{\quad}
\end{equation}
\begin{equation*}
	\vec X_4 = (0,0,0,1)
\end{equation*}
and the coefficients $k_1$ and $k_2$ represent a distance in the $(x,y,\theta)$ domain and $k_4$ is a Euclidean distance.
Starting from these vector fields we can model the cortical connectivity.

\subsection{The Connectivity Kernels}
\label{sec:connectivity}

The cortical connectivity can be modelled as the probability of connecting two points in the cortex and is represented by the stochastic counterpart of the curves in Eq. \ref{eq1}: 
\begin{equation}
	(x',y',\theta') = \vec X_1 + N(0,\sigma^2)\vec X_2 
	\label{eq:crticalConnectivity}
\end{equation}
where $N(0,\sigma^2)$ is a normally distributed variable with zero mean and variance equal to $\sigma^2$. 

This process, first described in \cite{mumford1994elastica}, is discussed in \cite{august2000curve,august2003sketches,sanguinetti2010model,williams1997stochastic}.
We denote $v$ the probability density to find a particle at the
point $(x, y)$ considering that it started from a given location $(x', y')$ and that it is moving
with some known velocity. This probability density
satisfies a deterministic equation known in literature as the
Kolmogorov forward equation or Fokker Planck equation:
\begin{equation*}
	\partial_tv=X_1v +\sigma^2X_{22}v
\end{equation*} 
where $X_1$ is the directional derivative $\cos(\theta)\partial_x + \sin(\theta)\partial_y$ and $X_{22}= \partial_{\theta\theta}$ is the second order derivative.

This equation has been largely used in different fields \cite{august2000curve,august2003sketches,williams1997stochastic,duits2010left}. In \cite{sanguinetti2010model} its stationary counterpart was proposed to model the probability of co-occurrence of contours in natural images.
The Fokker Planck operator has a nonnegative fundamental solution $\Gamma_1$ that satisfies: 
\begin{equation*}
\begin{split}
	X_1\Gamma_1((x,y,\theta),(x',y',\theta')) + & \sigma^2X_{22}\Gamma_1((x,y,\theta),(x',y',\theta'))
	 \\ & =\delta(x,y,\theta)
\end{split}
	\label{Gamma}
\end{equation*}
which is not symmetric. The connectivity kernel  $\omega_1$ obtained by symmetrization of the Fokker Planck fundamental solution is:
\begin{equation*}
	\begin{split}
	\omega_1((x,y,\theta),(x',y',\theta')) \\
	= \frac{1}{2}(\Gamma_1((x,y,\theta),(x',y',\theta')) +\Gamma_1((x',y',\theta'),(x,y,\theta)).
	\end{split}
	\label{omega}
\end{equation*}
In order to measure the distances between intensities we introduce the kernel $\omega_2((x_i,y_i),(x_j,y_j))$. This new intensity kernel is obtained as:
\begin{equation}
	\omega_2(f_i,f_j) = e^{(-\frac{1}{2}(\frac{f_i-f_j}{\sigma_2}))^2}
	\label{eq:omega2}
\end{equation} 

The final connectivity kernel can be written as the product (as these are probabilities) of the two components: 
\begin{equation}
\begin{split}
	\omega_f((x_i,y_i,\theta_i,f_i),(x_j,y_j,\theta_j,f_j)) \\ = \omega_1((x_i,y_i,\theta_i),(x_j,y_j,\theta_j))\omega_2(f_i,f_j). 
\end{split}
\label{eq:omegaf}
\end{equation} 


\subsection{Affinity Matrix}
\label{sec:affinity}

Starting from the connectivity kernel defined previously, it is possible to extract perceptual units from images by means of spectral analysis of suitable affinity matrices.  The eigenvectors with the highest eigenvalues are linked to the most salient objects in the scene \cite{perona1998factorization}. The connectivity is represented by a real symmetric matrix $A_{i,j}$:
\begin{equation} 
A_{i,j} = \omega_f((x_i,y_i,\theta_i,f_i), (x_j,y_j,\theta_j,f_j))
\label{eq:affinity}
\end{equation}
that contains the connectivity information between all the lifted points. The eigenvectors of the affinity matrix are interpreted as perceptual units \cite{sarti2015constitution}. 

\section {Spectral Analysis}
\label{sec:specClust}

The goal of clustering is to divide the data points into several groups such that points in the same group are similar and points in different groups are dissimilar to each other.
The cognitive task of visual grouping can be considered as a form of clustering, with which it is possible to separate points in different groups according to their similarities. In order to perform visual grouping, we will use the spectral clustering algorithm.
Traditional clustering algorithms, such as K-means, are not able to resolve this problem \cite{ng2002spectral}. In recent years, different techniques have been presented to overcome the performance of the traditional algorithms, in particular spectral analysis techniques. It is widely known that these techniques can be used for data partitioning and image segmentation \cite{shi2000normalized,perona1998factorization,weiss1999segmentation,meila2001random} and they outperform the traditional approaches. Above that, they are simple to implement and can be solved efficiently by standard linear algebra methods \cite{von2007tutorial}.
In the next section we will describe the spectral clustering algorithm used in the numerical simulations of this paper. 

\subsection{Spectral Clustering Technique}

Different algorithms based on the theory of graphs have been proposed to perform clustering. In \cite{perona1998factorization} it has been shown how the edge weights $\{$$a_{ij}$$\}$$_{i,j=1,...n}$ of a weighted graph describe an affinity matrix \textit{A}. This matrix contains information about the correct segmentation and will identify perceptual units in the scene, where the salient objects will correspond to the eigenvectors with the highest eigenvalues. Even though it works successfully in many examples, in \cite{weiss1999segmentation} it has been demonstrated that this algorithm also can lead to clustering errors.
In \cite{von2007tutorial} and \cite{meila2001random} the algorithm is improved considering the normalized affinity matrix. In particular we will use the normalization described in \cite{meila2001random}. Defining the diagonal matrix \textit{D} as formed by the sum of the edge weights (representing the degrees of the nodes,
$d_i = \sum_{j = 1}^{n} a_{ij}$), the normalized affinity matrix is obtained as:
\begin{equation}
	P = D^{-1}A
	\label{P}
\end{equation}
This stochastic matrix P represents the transition probability of a random walk in a graph. It has real eigenvalues $\{$$\lambda_{j}$$\}$$_{j=1,...n}$ where $0\leq\lambda_{j}\leq 1$, and its eigenvectors $\{$$u_{i}$$\}$$_{i=1,...K}$, related to the K largest eigenvalues $\lambda_{1}\geq\lambda_{2}\geq...\geq\lambda_{K}$, represent a solution of the clustering problem \cite{von2007tutorial}. 
The value of K determines the number of eigenvalues and eigenvectors considered informative. 
 
\begin{figure*}[htbp!]
	\centering
	\begin{subfigure}[b]{1.6 in}
		\includegraphics[width= \columnwidth]{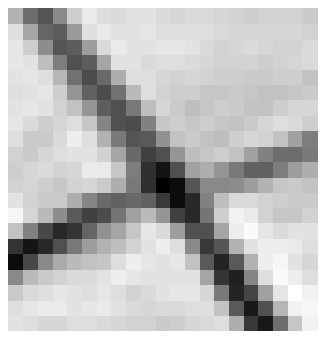} \caption{Image patch}
		\label{fig:den2}
	\end{subfigure}
	\begin{subfigure}[b]{1.6 in}
		\includegraphics[]{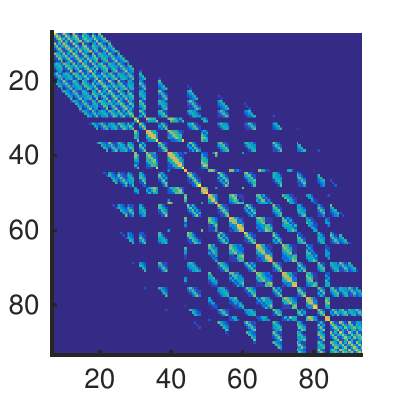} \caption{Affinity matrix}
		\label{fig:aff2}
	\end{subfigure}
	\begin{subfigure}[b]{1.6 in}
		\includegraphics{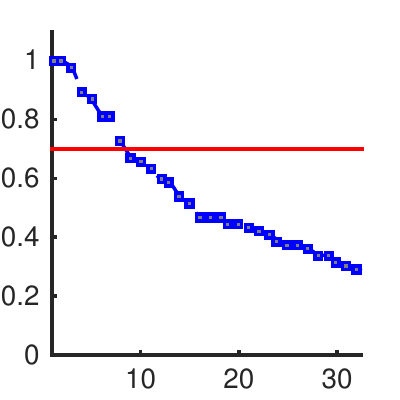}  \caption{Eigenvalues}
		\label{fig:eig}
	\end{subfigure}
	\begin{subfigure}[b]{1.6 in}		
		\includegraphics{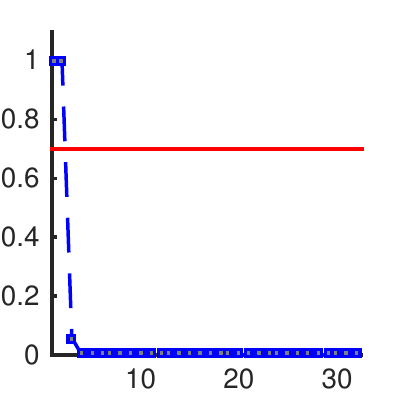}  \caption{Eigenvalues, $\tau$ = 150}
		\label{fig:eigtau} 
	\end{subfigure}
	\caption{
		(\subref{fig:den2}) A sample image patch at a crossing, (\subref{fig:aff2}) its affinity matrix ($A$) built upon a connectivity measure, (\subref{fig:eig}) the eigenvalues ($\lambda_i,{i=1,...n}$) of the normalized affinity matrix ($P$) and the threshold value ($1-\epsilon=0.7$), represented in red. (\subref{fig:eigtau}) The exponentiated spectrum ($\lambda^\tau_i,{i=1,...n}$) with $\tau=150$ and threshold value of $0.7$ in red.
%
		}
	\label{fig:eigenvalues}
\end{figure*}

The important step is selecting the best value of K, which can be done by defining an a-priori significance threshold $\epsilon$ for the decreasingly ordered eigenvalues $\lambda_{i}$, so that $\lambda_{i} > 1-\epsilon, \forall~1 \leq i \leq K$. However, selecting the best $\epsilon$ value is not always trivial, and the clustering results get very sensitive to this parameter in many cases. 
Hence, 
considering the diffusion map approach of~\cite{coifman2006diffusion} and following the idea of~\cite{cocci2015cortical}, using an auxiliary diffusion parameter ($\tau$, big positive integer value) to obtain the exponentiated spectrum $\{\lambda_{i}^\tau\}_{i=1,\dots n}$, the gap between exponentiated eigenvalues increases and sensitivity to the threshold value decreases very much. 
Using this new spectrum, yields to the stochastic matrix $P^\tau$, that represents the transition matrix of a random walk in defined $\tau$ steps. The difference between thresholding the eigenvalues directly or the exponentiated spectrum is shown in an example in Fig.~\ref{fig:eigenvalues}. As seen in this figure, selecting the best discriminative threshold value for the eigenvectors (Fig.~\ref{fig:eig}) is not easy, while with the exponentiated spectrum (Fig.~\ref{fig:eigtau}) the threshold value can be selected in a wide range (e.g. $0.05\leq 1-\epsilon \leq  0.9$). The value of $\tau$ need to be selected as a large positive integer number (e.g. $150$). 

After selecting the value of K, the number of clusters is automatically determined using Algorithm~\ref{alg:spec}. 
\begin{algorithm}
\caption{Spectral clustering algorithm} 
\begin{algorithmic} [1]
\label{alg:spec}
  \STATE Define the affinity matrix $A_{i,j}$ from the connectivity kernel.
  \STATE Evaluate the normalized affinity matrix: $P = D^{-1}A.$
  \STATE Solve the eigenvalue problem $Pu_i = \lambda_i u_i$, where the order of $i$ is such that $\lambda_i$ is decreasing.
  \STATE Define the thresholds $\epsilon$, $\tau$ and evaluate the largest integer K such that  $\lambda_i^\tau>1-\epsilon$, $i = 1,\dots,K$.
  \STATE Let U be the matrix containing the vectors $u_1,\dots,u_K$ as columns.
  \STATE Define the clusters k = $\argmax_j \{u_j(i)\}$ with $j\in \{1,\dots,K\}$ and $i=1,\dots,n$.
  \STATE Find and remove the clusters that contain less than a minimum cluster size elements.
\end{algorithmic}
\end{algorithm}

Possible neural implementations of the algorithm are discussed in \cite{cocci2015cortical}. Particularly, in \cite{bressloff2002geometric,faugeras2009persistent} an implementation of the spectral analysis is described as a mean-field neural computation. Principal eigenvectors emerge as symmetry breaking of the stationary solutions of mean field equations. In addition, in \cite{sarti2015constitution} it is shown that in the presence of a visual stimulus the emerging
eigenvectors are linked to visual perceptual units, obtained from a
spectral clustering on excited connectivity kernels.
In the next section the application of this algorithm in obtaining the vessel clustering in retinal images will be presented.

\section{Experiments and Results}
\label{sec:experiments}
In this section, the steps proposed for analyzing the connectivities of blood vessels in retinal images and validating the method are described. In addition, the parameter settings and the obtained results are discussed in detail. 
\subsection{Proposed Technique}
\begin{figure*}[htbp]
  \centering 
 	 \begin{subfigure}[b]{0.3\textwidth}
        		 \includegraphics[width=1  \textwidth]{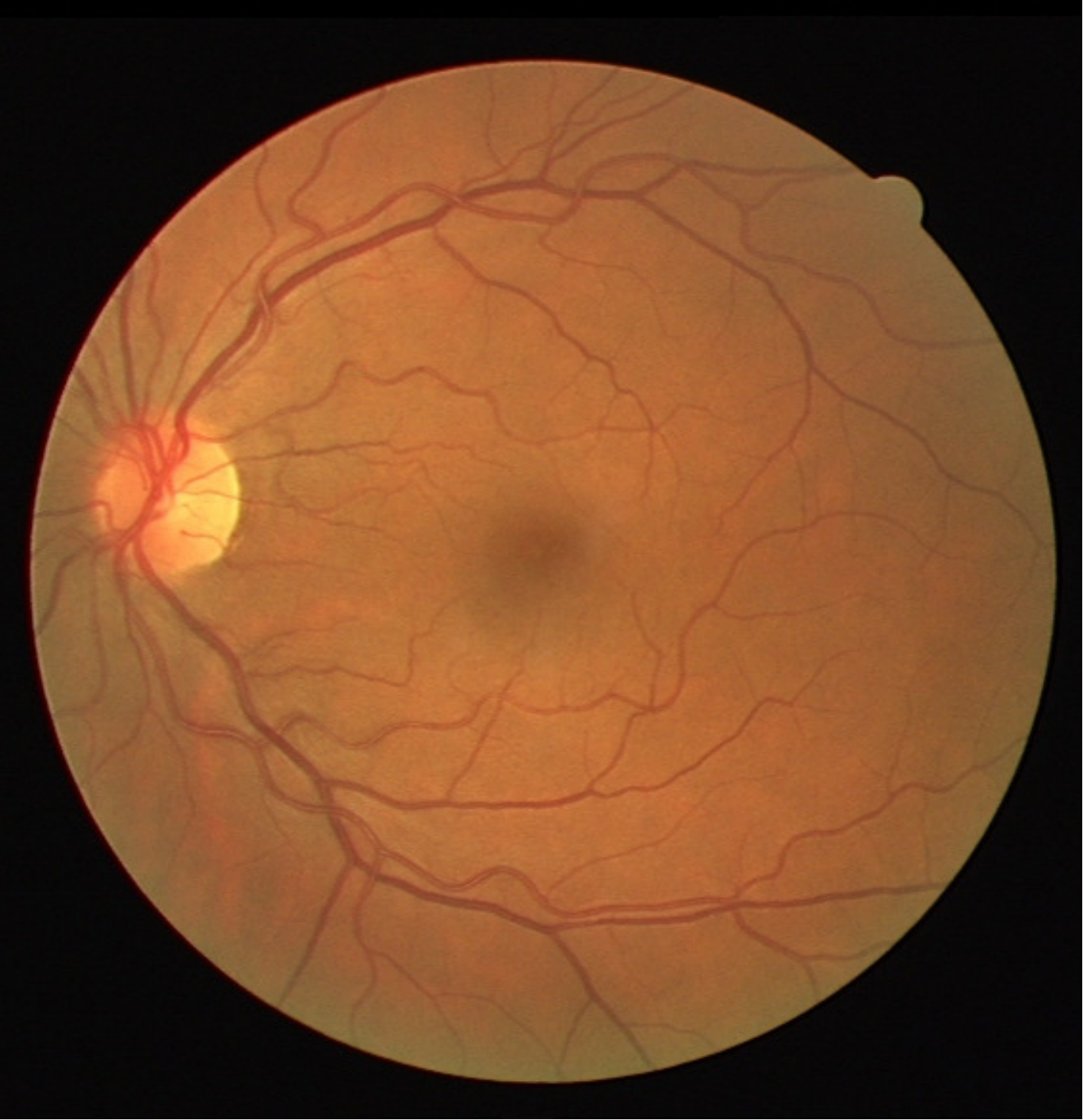} \caption{} \label{fig:colored}
        	\end{subfigure}
	\begin{subfigure}[b]{0.3\textwidth}
        		 \includegraphics[width= 1\textwidth]{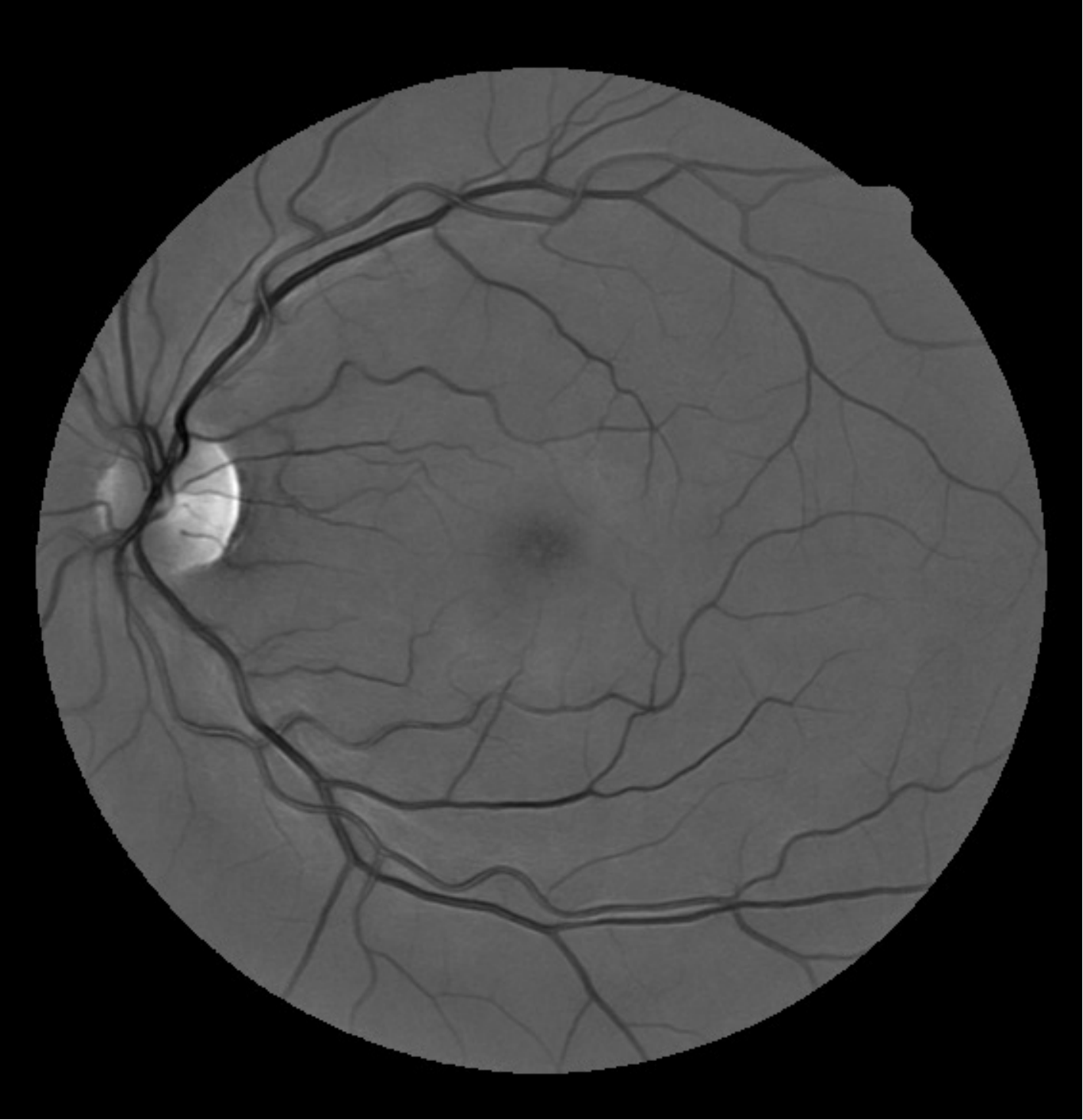} \caption{} \label{fig:imEnhanced}
        	\end{subfigure}
	\begin{subfigure}[b]{0.3 \textwidth}
        		 \includegraphics[width= 1\textwidth]{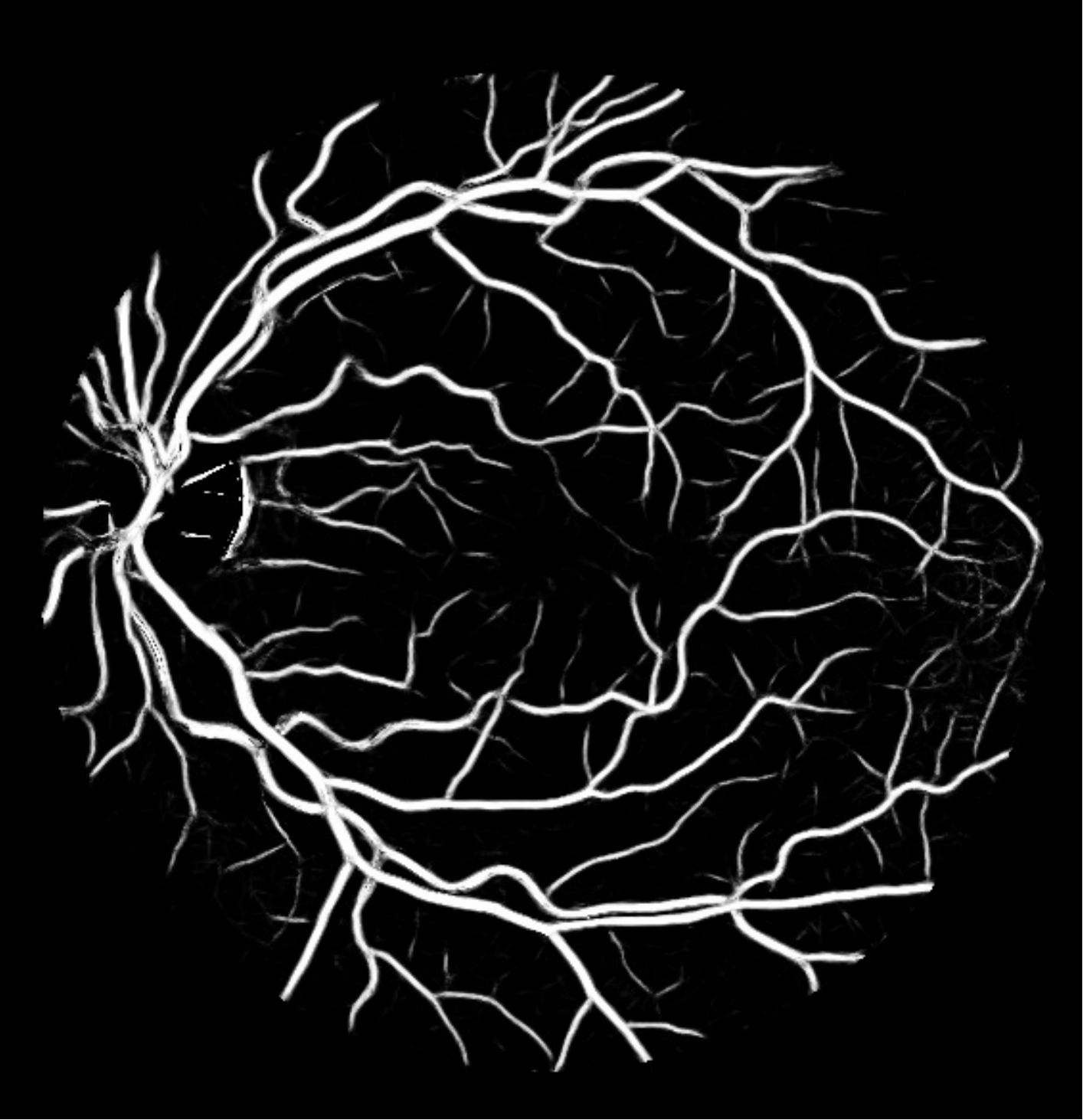} \caption{} \label{fig:BIMSOsoftSeg}
        	\end{subfigure}\\
	\begin{subfigure}[b]{0.3\textwidth}
        		 \includegraphics[width=1 \textwidth]{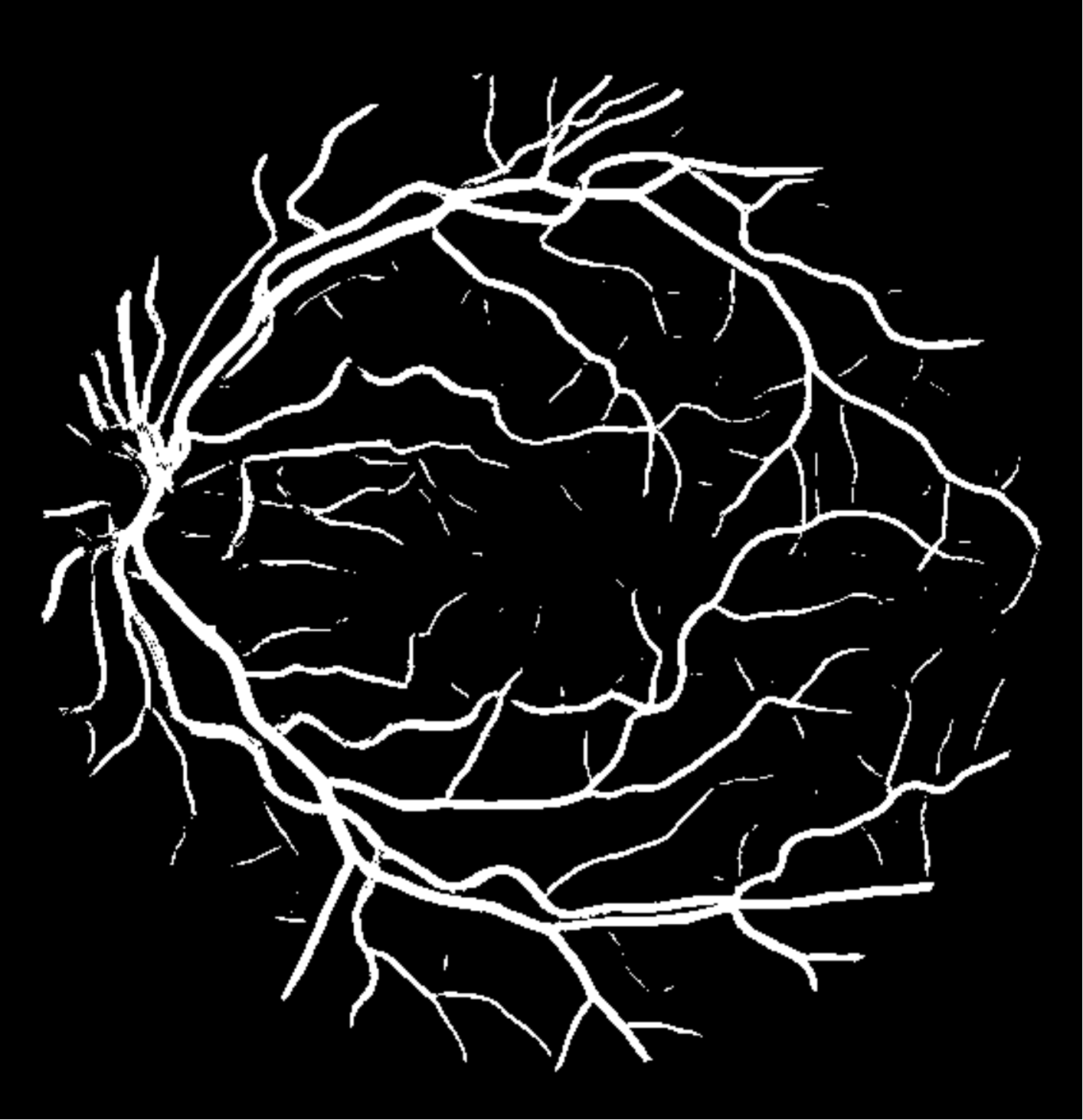} \caption{} \label{fig:BIMSOhardSeg}
        	\end{subfigure}
	\begin{subfigure}[b]{0.3\textwidth}
	 \includegraphics[width=1 \textwidth]{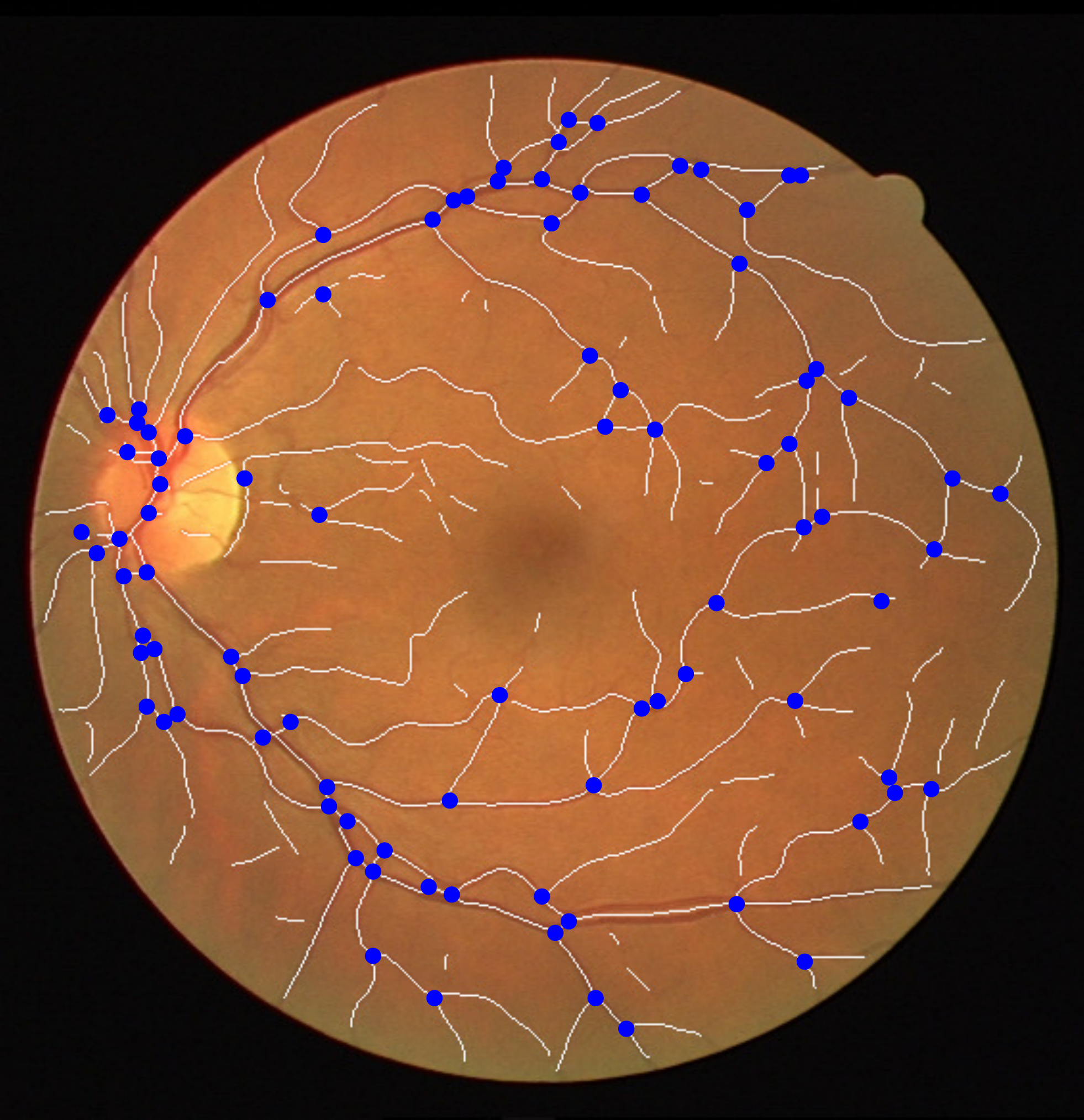}\caption{} \label{fig:junctions}
	 \end{subfigure}
 \begin{subfigure}[b]{0.3\textwidth}
  \includegraphics[width= 1\textwidth]{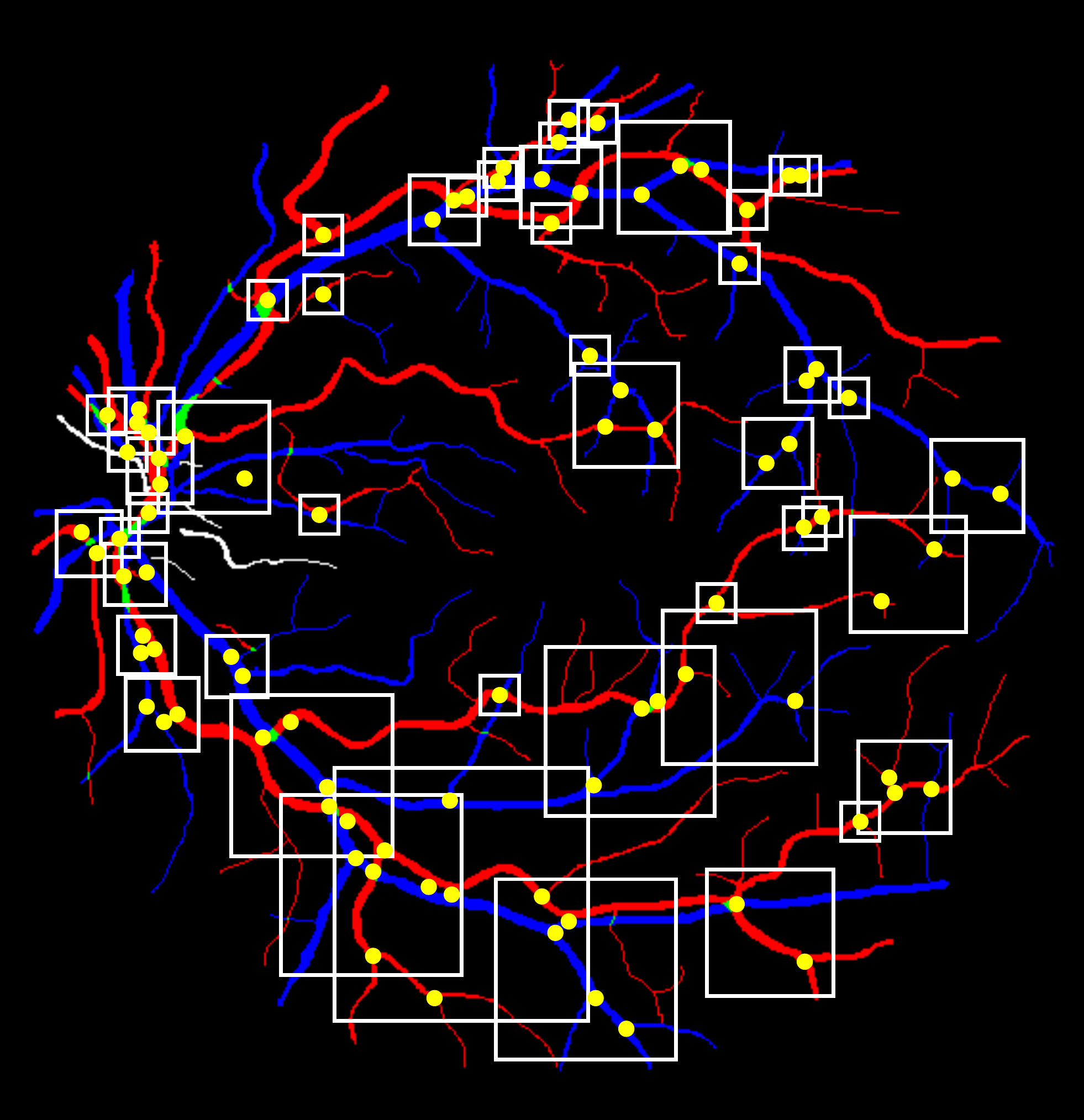}\caption{}\label{fig:imagePatches}
  \end{subfigure}
  \caption{The different steps applied for selecting several image patches around junctions, (\subref{fig:colored}) original RGB image, (\subref{fig:imEnhanced}) enhanced image ($I_{enh}$), (\subref{fig:BIMSOsoftSeg}) soft segmentation ($I_{soft}$), (\subref{fig:BIMSOhardSeg}) hard segmentation ($I_{hard}$), (\subref{fig:junctions}) detected junctions and the skeleton of the segmentation overlaid on color image, (\subref{fig:imagePatches}) selected patches overlaid on artery/vein ground truth }
  \label{fig:steps4imagePatchSelection}
\end{figure*}

In order to prove the reliability of the method in retrieving the connectivity information in 2D retinal images, several challenging and problematic image patches around junctions were selected. First step before detecting the junctions and selecting the image patches around them, is to apply preconditioning on the green channel ($I$) of a color fundus retinal image. The green channel provides a higher contrast between vessels and background and it is widely used in retinal image analysis. The preconditioning includes: 
\begin{inparaenum}[\itshape a\upshape)]
\item removing the non-uniform luminosity and contrast variability using the method proposed by \cite{Foracchia2005};
\item removing the high frequency contents; and 
\item denoising using the non-linear enhancement in SE(2) as proposed by \cite{abbasi2015biologically}
\end{inparaenum}. A sample color image before and after preconditioning ($I_{enh}$) are shown in Fig.~\ref{fig:colored} and~\ref{fig:imEnhanced} respectively.

In next step, soft ($I_{soft}$) and hard ($I_{hard}$) segmentations are obtained using the BIMSO (biologically-inspired multi-scale and multi-orientation) method for segmenting $I_{enh}$ as proposed by~\cite{abbasi2015biologically}. These images are shown in Fig.~\ref{fig:BIMSOsoftSeg} and~\ref{fig:BIMSOhardSeg} respectively. The hard segmentation is used for detecting the junctions and selecting several patches with different sizes around them; while soft segmentation is used later in connectivity analysis. 

In order to find the junction locations automatically, the skeleton of $I_{hard}$ is produced using the morphological skeletonization technique. Then the method proposed by~\cite{olsen2011convolution} is applied on this skeleton and the junction locations are determined as shown in Fig.~\ref{fig:junctions}. Using the determined locations, several image patches with similar sizes ($s=10$ pixels) are selected at first stage. However, as seen in  Fig.~\ref{fig:junctions}, some of the junctions are very close to each other and their distances are smaller than $s/2$. For these junctions, a new patch including both nearby junctions (with a size equal to three times the distance between them) is considered, and its centre is used for finding the distance of this new patch with the other ones. These steps are repeated until no more merging is possible or the patch size reaches the maximum possible size (we assumed 100 as the maximum possible value). Thus, all nearby junctions are grouped in order to decrease the number of patches that overlap in a great extent. %
%
%
%
This results in having different patch sizes ($0 \leq s_{p_{i}}\leq 100,1\leq i\leq m$) that could include more than one junction all over the image. Fig.~\ref{fig:imagePatches} shows the junction locations and the corresponding selected patches overlaying on artery/vein ground truth. 

In order to analyze the vessel connectivities for each image patch ($I_{p_i}$), we need to extract the location ($x,y$), orientation ($\theta$) and intensity ($f(x,y)$) of vessel pixels in these patches. Hence for each group of junctions ($i$) with the size $s_i$, two patches from $I_{enh}$ and $I_{soft}$ are selected, called $I_{enh,p_i}$ and $I_{soft,p_i}$ respectively.
Then $I_{soft,p_i}$ is thresholded locally to obtain a new hard segmented image patch (called $I_{hard,p_i}$). This new segmented image patch is different from selecting the corresponding patch from $I_{hard}$, because $I_{hard}$
%
%
was obtained by thresholding the entire $I_{soft}$ using one global threshold value, but this is not appropriate at all regions. If there are regions with very small vessels with a low contrast (often they get a very low probability of being vessel pixels), they are normally removed in the global thresholding approach. 
Accordingly, wrong thresholding leads into wrong tracking results e.g. C1, C2, C6 in Fig. \ref{fig:trackingIssues} are some instance patches with missing small vessels. In this work, we selected one threshold value for each patch specifically using Otsu's method~\cite{otsu1975threshold}, to keep more information and cover a wider range of vessel pixels.  Consequently, thicker vessels will be created in $I_{hard,p_i}$ and the results will be more accurate.

By knowing the vessel locations $(x,y)$ other information could be extracted for these locations using $I_{enh,p_i}$. So $f(x,y)$ equals the intensity value in $I_{enh,p_i}$ at location $(x,y)$. 
Moreover, by lifting $I_{enh,p_i}$ using cake wavelets (see Eq.~\ref{eq:OST}), at each location the angle corresponding to the maximum of the negative orientation response (real part) in the lifted domain is considered as the dominant orientation ($\theta_d$) as Eq.~\ref{eq:MIP}. The negative response is considered because the blood vessels in retinal images are darker than background. 
\begin{equation} 
 \theta_d = \argmax_{\theta\in[0,\pi]} Re(-U_f(x,y,\theta))
 \label{eq:MIP}
 \end{equation}

Next step is approximating the connectivity kernels. The first kernel ($\omega_1$), was calculated numerically. So the fundamental solution $\Gamma_1$ was estimated using the Markov Chain Monte Carlo method~\cite{robert2013monte} by developing random paths based on the numerical solution of Eq.~\ref{eq:crticalConnectivity}. 
This solution can be approximated by:
\begin{equation*}
\begin{cases} 
x_{s+\Delta{s}}-x_s = \Delta{s}\cos(\theta)\\ 
y_{s+\Delta{s}}-y_s = \Delta{s}\sin(\theta)\\
\theta_{s+\Delta{s}} - \theta_s=\Delta{s}N(0,\sigma)
 \end{cases}, s\in 0, \dots, H
\end{equation*}where $H$ is the number of steps of the random path and $\sigma$ is the diffusion constant (the propagation variance in the $\theta$ direction). This finite difference equation is solved for $n$ (typically $10^5$) times, so $n$ paths are created. Then the estimated kernel is obtained by averaging all the solutions~\cite{higham2001algorithmic,sarti2015constitution}. An overview of different possible numerical methods to compute the kernel is explained in \cite{zhang2014numerical}, where comparisons are done with the exact solutions derived in \cite{duits2010left,DuitsCASA2005,duits2008explicit}.
From these comparisons it follows that the stochastic Monte-Carlo implementation is a fair and accurate method. The intensity-based kernel ($\omega_2$), the final connectivity kernel ($\omega_f$) and the affinity matrix ($A$), were calculated using Eq.~\ref{eq:omega2}, ~\ref{eq:omegaf} and \ref{eq:affinity} respectively. Finally, by applying the proposed spectral clustering step in Sect.~\ref{sec:specClust}, the final perceptual units (individual vessels) were obtained for each patch. 
		 		 
\begin{figure*}[htbp]
\centering
	\begin{subfigure}[b]{0.91 in}
	 \includegraphics[width= \columnwidth]{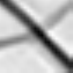}\caption{}\label{fig:sampledenoised}
	 \end{subfigure}
	 \begin{subfigure}[b]{0.91 in}
	 \includegraphics[width=  \columnwidth]{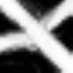}\caption{}\label{fig:samplesoftSeg}
	 \end{subfigure}
	 \begin{subfigure}[b]{0.91 in}
	 \includegraphics[width=\columnwidth]{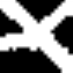}\caption{}\label{fig:samplehardSeg}
	 \end{subfigure}
	  \begin{subfigure}[b]{0.91 in}
	 \includegraphics[width=\columnwidth]{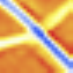}\caption{}\label{fig:sampleIntensity}
	 \end{subfigure}
	 \begin{subfigure}[b]{0.91 in}
	 \includegraphics[width=\columnwidth]{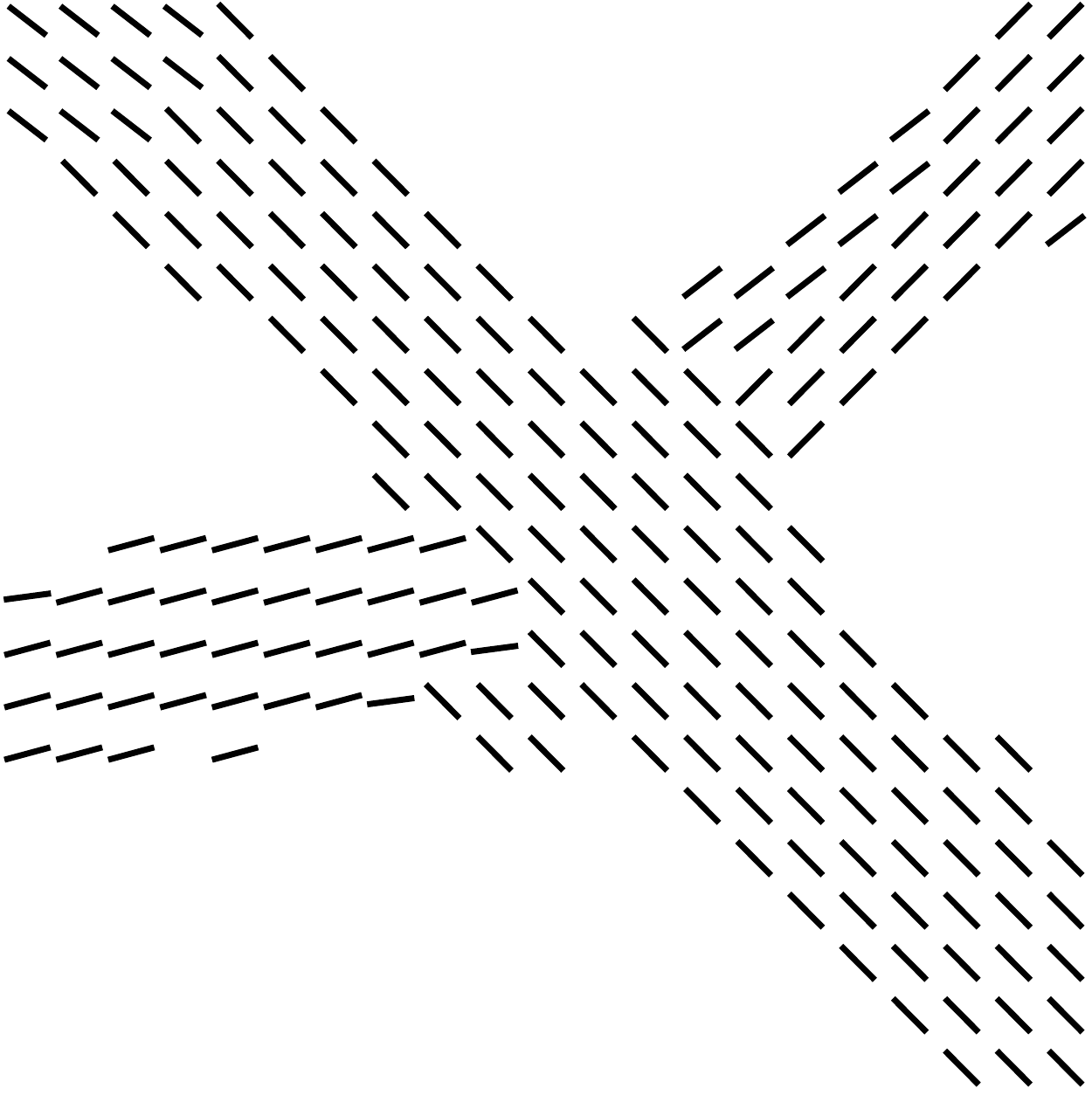}\caption{}\label{fig:sampleOrientation}
	 \end{subfigure}
	  \begin{subfigure}[b]{0.91 in}
	 \includegraphics[width=\columnwidth]{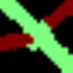}\caption{}\label{fig:sampleResult2}
	 \end{subfigure}
	 \begin{subfigure}[b]{0.91 in}
	 \includegraphics[width=\columnwidth]{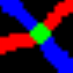}\caption{}\label{fig:sampleavGT}
	 \end{subfigure}
	 
	 \begin{subfigure}[b]{1.6 in}
	 \includegraphics{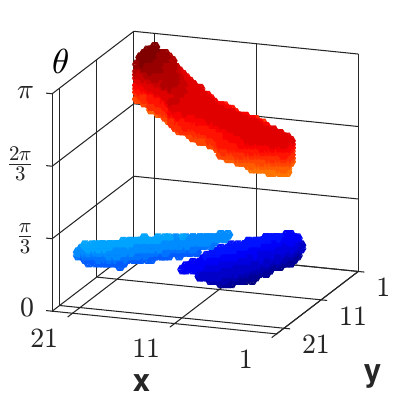}\caption{}\label{fig:sampleOSprofile}
	 \end{subfigure}
	  \begin{subfigure}[b]{1.6 in}
	 \includegraphics{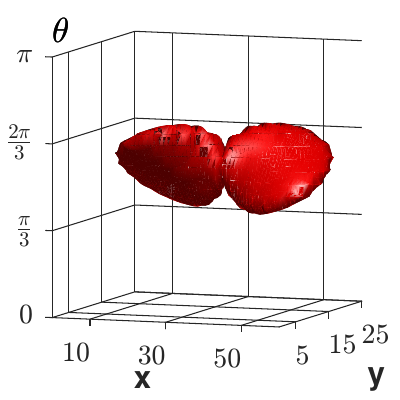}\caption{}\label{fig:sampleKernel}
	 \end{subfigure}
	 \begin{subfigure}[b]{1.6 in}
	 \includegraphics{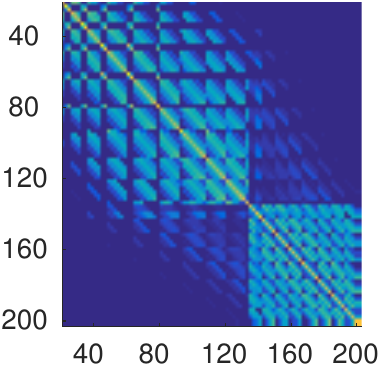}\caption{}\label{fig:sampleAffinity}
	 \end{subfigure}
	  \begin{subfigure}[b]{1.6 in}
	 \includegraphics{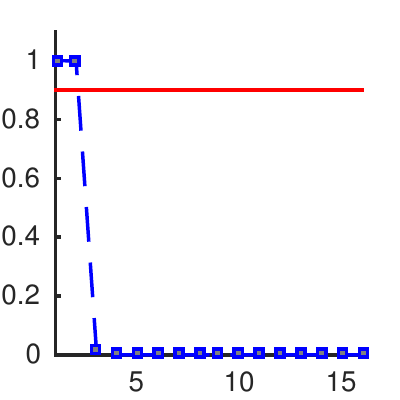}\caption{}\label{fig:sampleEigenValues}
	 \end{subfigure}
	 
	 \caption{A sample $21\times 21$ image patch at a crossing: (\subref{fig:sampledenoised}) $I_{enh,p_i}$, (\subref{fig:samplesoftSeg}) $I_{soft,p_i}$, (\subref{fig:samplehardSeg}) $I_{hard,p_i}$, (\subref{fig:sampleIntensity}) the differences in intensities are shown in color, (\subref{fig:sampleOrientation}) each oriented line represents the orientation at its position, (\subref{fig:sampleResult2}) final perceptual units shown in different colors (\subref{fig:sampleavGT}) the ground truth artery and vein labels, (\subref{fig:sampleOSprofile}) lifted image in  $SE(2)$, (\subref{fig:sampleKernel}) connectivity kernel ($\omega_2$), (\subref{fig:sampleAffinity}) affinity matrix ($A$) obtained using both orientation and intensity information, (\subref{fig:sampleEigenValues}) thresholding the eigenvalues of the normalized affinity matrix}\label{fig:Clusteringsteps}
\end{figure*}

The above-mentioned steps for a sample crossing in a $21\times 21$ image patch are presented in Fig.~\ref{fig:Clusteringsteps}. After enhancing the image (Fig. \ref{fig:sampledenoised}), obtaining soft segmentation (Fig.~\ref{fig:samplesoftSeg}) and thresholding it locally (Fig.~\ref{fig:samplehardSeg}), the vessel locations, intensity and orientation have been extracted. As shown in Fig.~\ref{fig:sampleIntensity} arteries and veins have different intensities and this difference helps in discriminating between them. Though, orientation information is the most discriminative one. 
The lifted image in $SE(2)$ using the $\pi$-periodic cake wavelets in $24$ different orientations is shown in Fig.~\ref{fig:sampleOSprofile}. The disentanglement of two crossing vessels at the junction point can be seen clearly in this figure.  The dominant orientations ($\theta_d$) for the vessel pixels are also depicted in Fig.~\ref{fig:sampleOrientation}, using line segments oriented according to the corresponding orientation at each pixel.

In the next step, this contextual information (intensity and orientation) is used for calculating the connectivity kernel (Fig.~\ref{fig:sampleKernel}) and the affinity matrix (Fig.~\ref{fig:sampleAffinity}) as mentioned in Sect.~\ref{sec:geometryV1}. For this numerical simulation, $H$, $n$, $\sigma$ and $\sigma_2$ have been set to $7$, $100000$, $0.05$ and $0.1$ respectively. 
Next, by applying the spectral clustering on the normalized affinity matrix using $\epsilon$ and $\tau$ as $0.1$ and $150$, only two eigenvalues above the threshold will remain (Fig.~\ref{fig:sampleEigenValues}). 
This means that there are two main salient perceptual units in this image as it was expected. These two units are color coded in Fig.~\ref{fig:sampleResult2}. The corresponding artery and vein labels are also depicted in Fig~\ref{fig:sampleavGT} which approve the correctness of the obtained clustering results. 

\subsection{Validation}
To validate the method, the proposed steps were applied on several image patches of the DRIVE~\cite{staal2004} dataset. This public dataset contains 40 color images with a resolution of $565\times 584$ ($\sim25\mu{m}/px$) and a $45^\circ$ field of view. The selected patches from each image were manually categorized into the following groups: simple crossing (category A), simple bifurcation (category B), nearby parallel vessels with bifurcation (category C), bifurcation next to a crossing (category D) and multiple bifurcations (category E), and each category narrowed down to 20 image patches. These patches have different complexities, number of junctions and sizes and they could contain broken lines, missing small vessels and vessels with high curvature. The parameters used in the numerical simulation of the affinity matrix and spectral clustering step (including $\sigma$, $H$, $n$, $\sigma$, $\sigma_2$, $\epsilon$ and $\tau$) are chosen for each patch differently, with the aim of achieving the optimal results for each case. Automatic parameter selection remains a challenging task and will be investigated in future work.

\begin{figure*}[htbp]
        \centering
        \begin{subfigure}[b]{0.9 in}\centering
		 \makebox[0pt][r]{\makebox[15pt]{\raisebox{25pt}{\rotatebox[origin=c]{90}{G1, A}}}}%
		 \includegraphics[width=\textwidth]{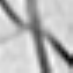} 
		 \makebox[0pt][r]{\makebox[15pt]{\raisebox{25pt}{\rotatebox[origin=c]{90}{G2, B}}}}%
		 \includegraphics[width=\textwidth]{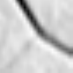} 
		 \makebox[0pt][r]{\makebox[15pt]{\raisebox{25pt}{\rotatebox[origin=c]{90}{G3, C}}}}%
		 \includegraphics[width=\textwidth]{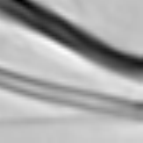}
		 \makebox[0pt][r]{\makebox[15pt]{\raisebox{25pt}{\rotatebox[origin=c]{90}{G4, E}}}}%
		 \includegraphics[width=\textwidth]{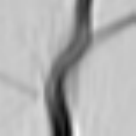}
		 \makebox[0pt][r]{\makebox[15pt]{\raisebox{25pt}{\rotatebox[origin=c]{90}{G5, E}}}}%
		 \includegraphics[width=\textwidth]{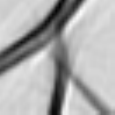}
		 \makebox[0pt][r]{\makebox[15pt]{\raisebox{25pt}{\rotatebox[origin=c]{90}{G6, D}}}}%
		 \includegraphics[width=\textwidth]{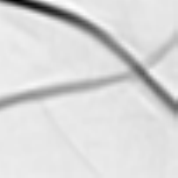} 
		 \makebox[0pt][r]{\makebox[15pt]{\raisebox{25pt}{\rotatebox[origin=c]{90}{G7, D}}}}%
		 \includegraphics[width=\textwidth]{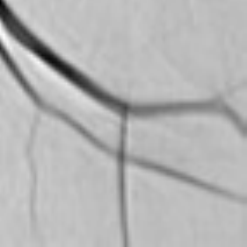}
		 \makebox[0pt][r]{\makebox[15pt]{\raisebox{25pt}{\rotatebox[origin=c]{90}{G8, D}}}}%
		 \includegraphics[width=\textwidth]{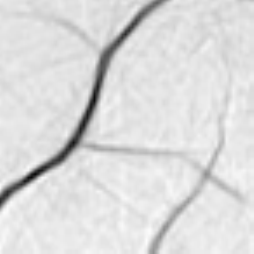}
		 \makebox[0pt][r]{\makebox[15pt]{\raisebox{25pt}{\rotatebox[origin=c]{90}{G9, C}}}}%
		 \includegraphics[width=\textwidth]{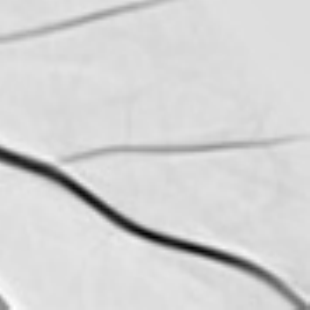}
		 \makebox[0pt][r]{\makebox[15pt]{\raisebox{25pt}{\rotatebox[origin=c]{90}{G10, D}}}}%
		 \includegraphics[width=\textwidth]{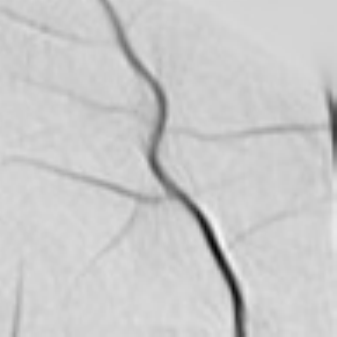}
		  \caption{}\label{fig:denoisedImg}
        \end{subfigure}
         \begin{subfigure}[b]{0.9 in}
         \centering
		 \includegraphics[width=\textwidth]{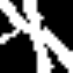}
		 \includegraphics[width=\textwidth]{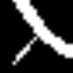}
		 \includegraphics[width=\textwidth]{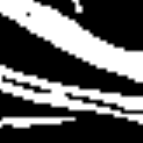}
		 \includegraphics[width=\textwidth]{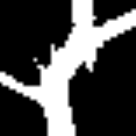}
		 \includegraphics[width=\textwidth]{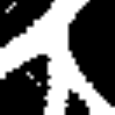} 
		 \includegraphics[width=\textwidth]{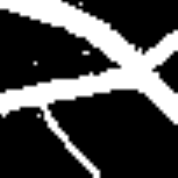}
		 \includegraphics[width=\textwidth]{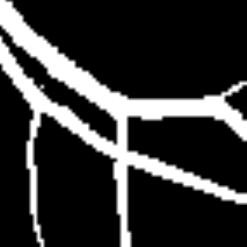}
		 \includegraphics[width=\textwidth]{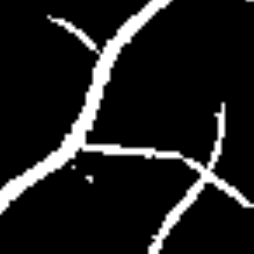}
		 \includegraphics[width=\textwidth]{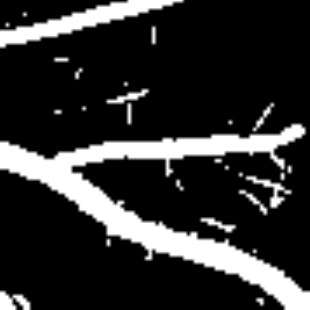} 
		 \includegraphics[width=\textwidth]{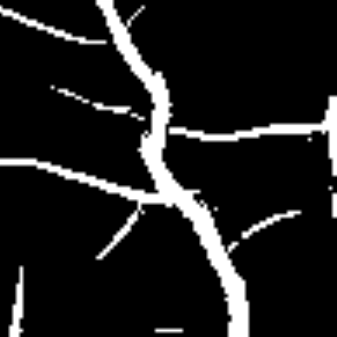}
		 \caption{}\label{fig:softSegThr}
        \end{subfigure}
         \begin{subfigure}[b]{0.9 in}
         \centering
		  \includegraphics[width=\textwidth]{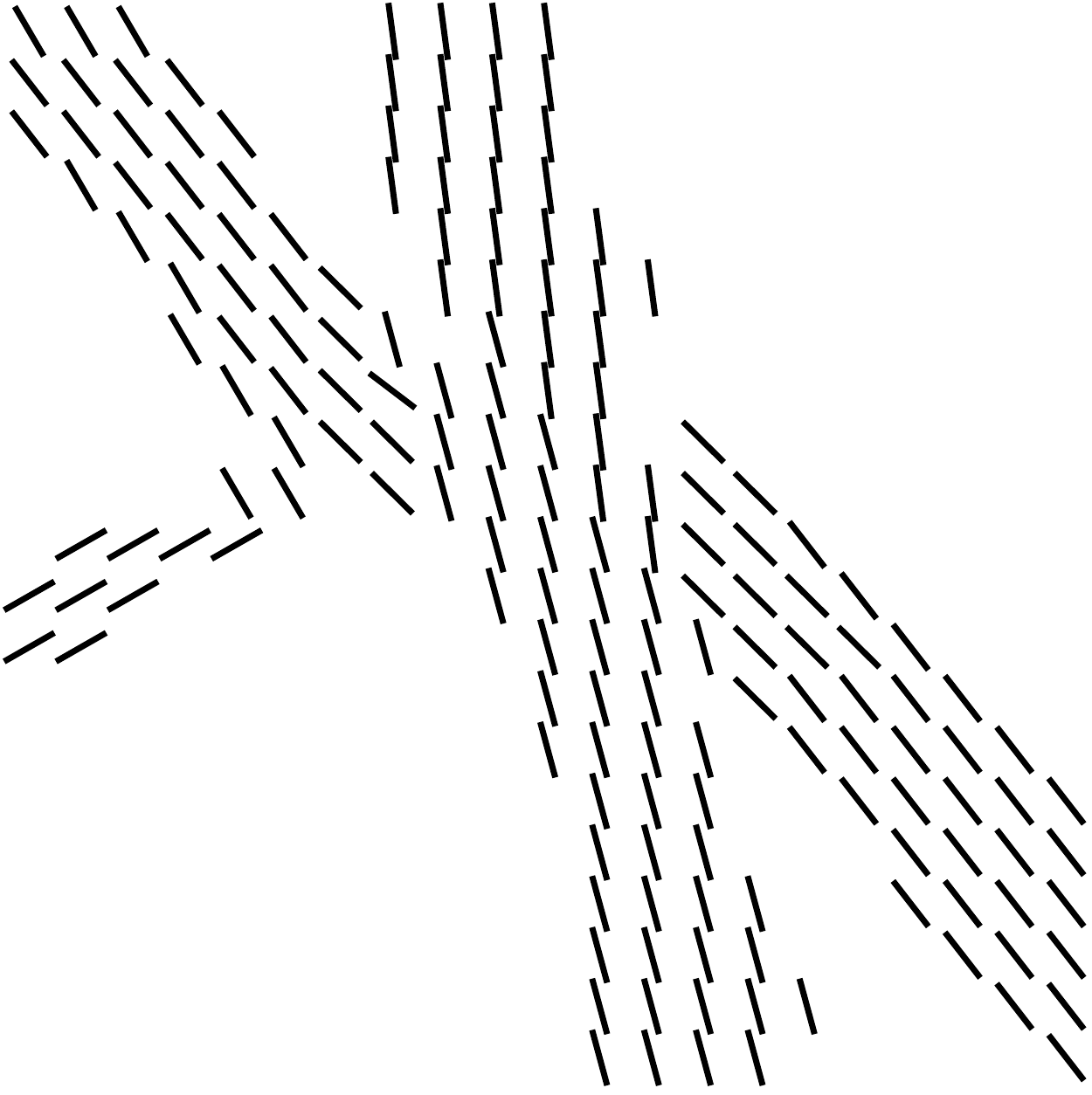}
		 \includegraphics[width=\textwidth]{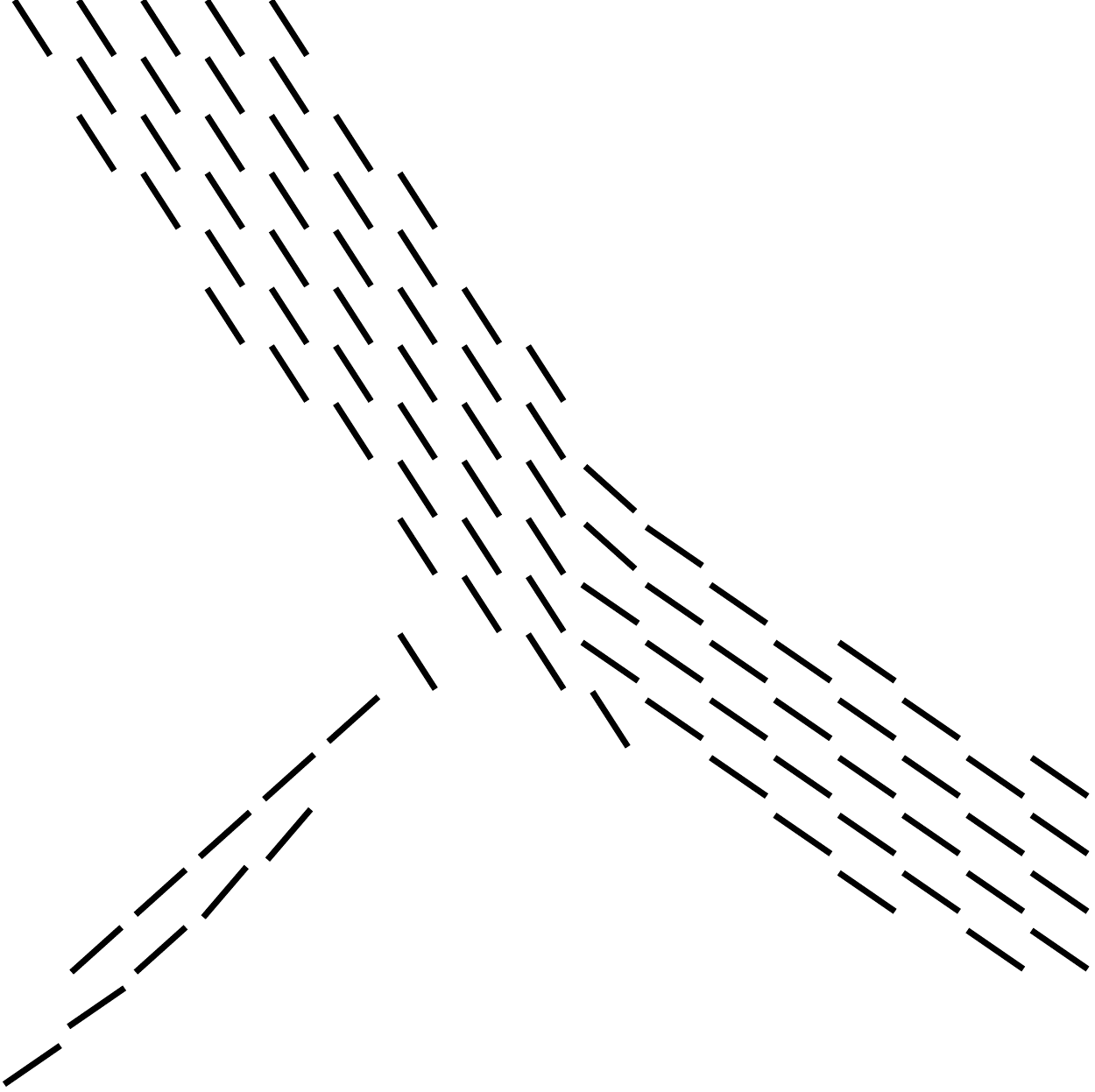}
		 \includegraphics[width=\textwidth]{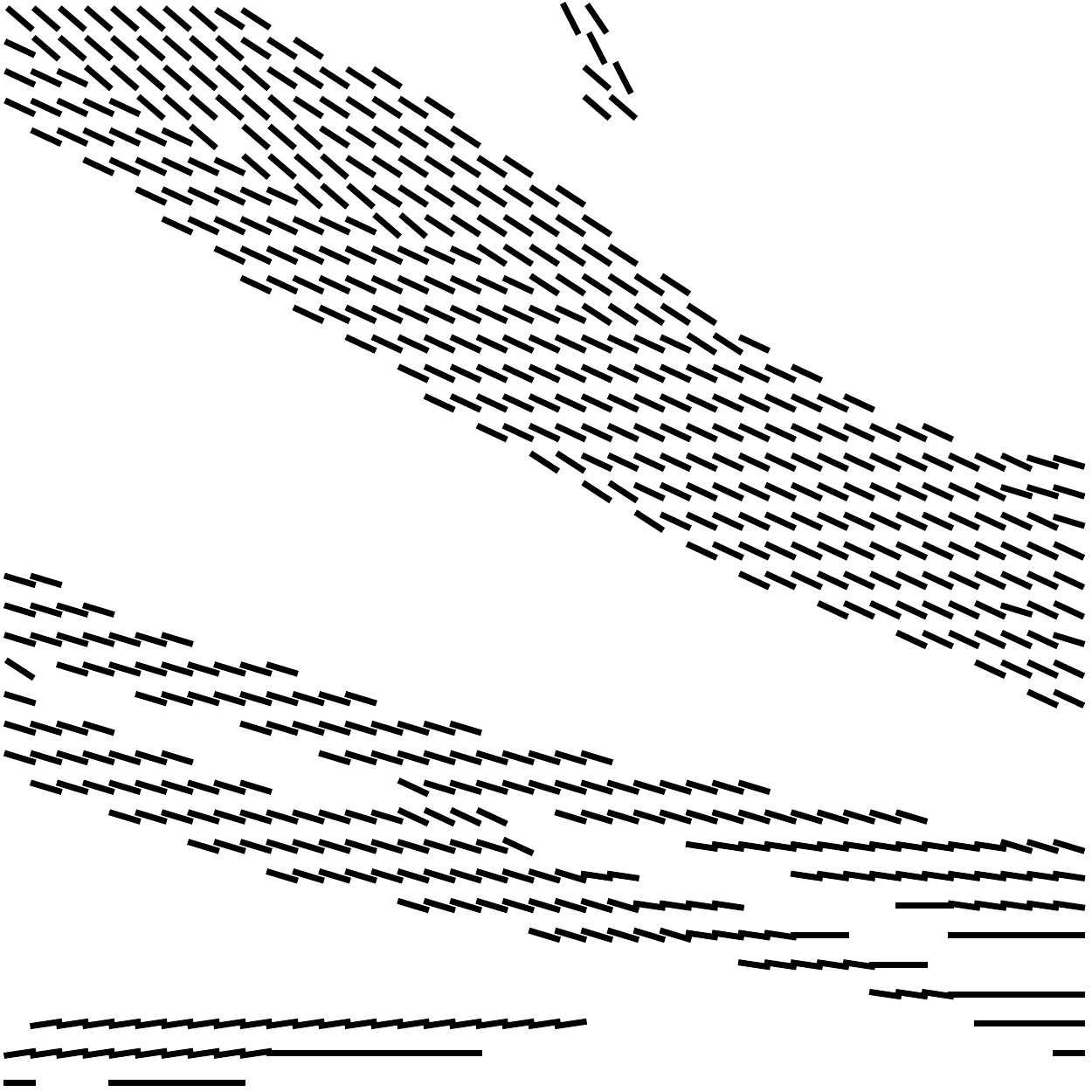}
		 \includegraphics[width=\textwidth]{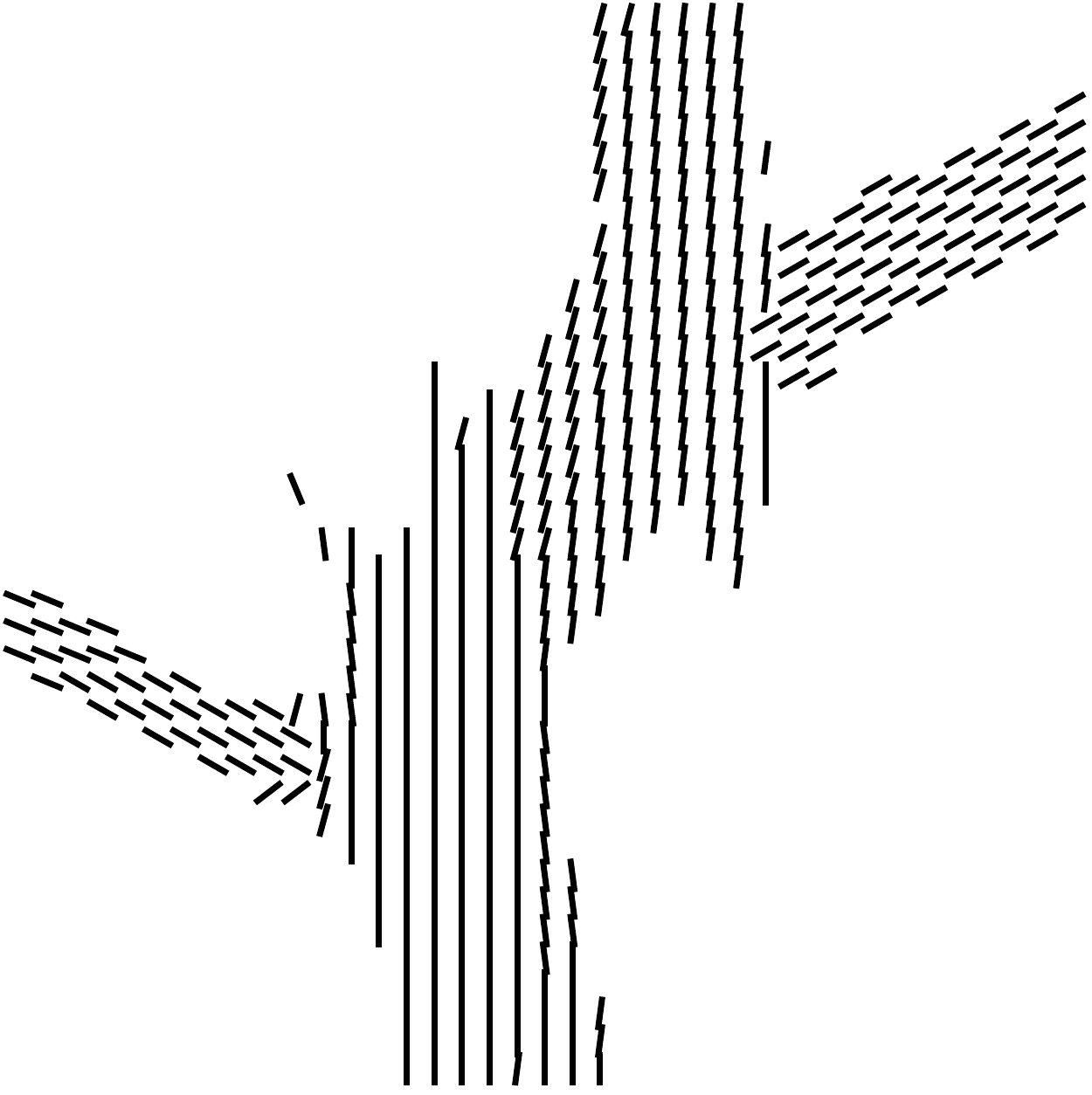}
        		 \includegraphics[width=\textwidth]{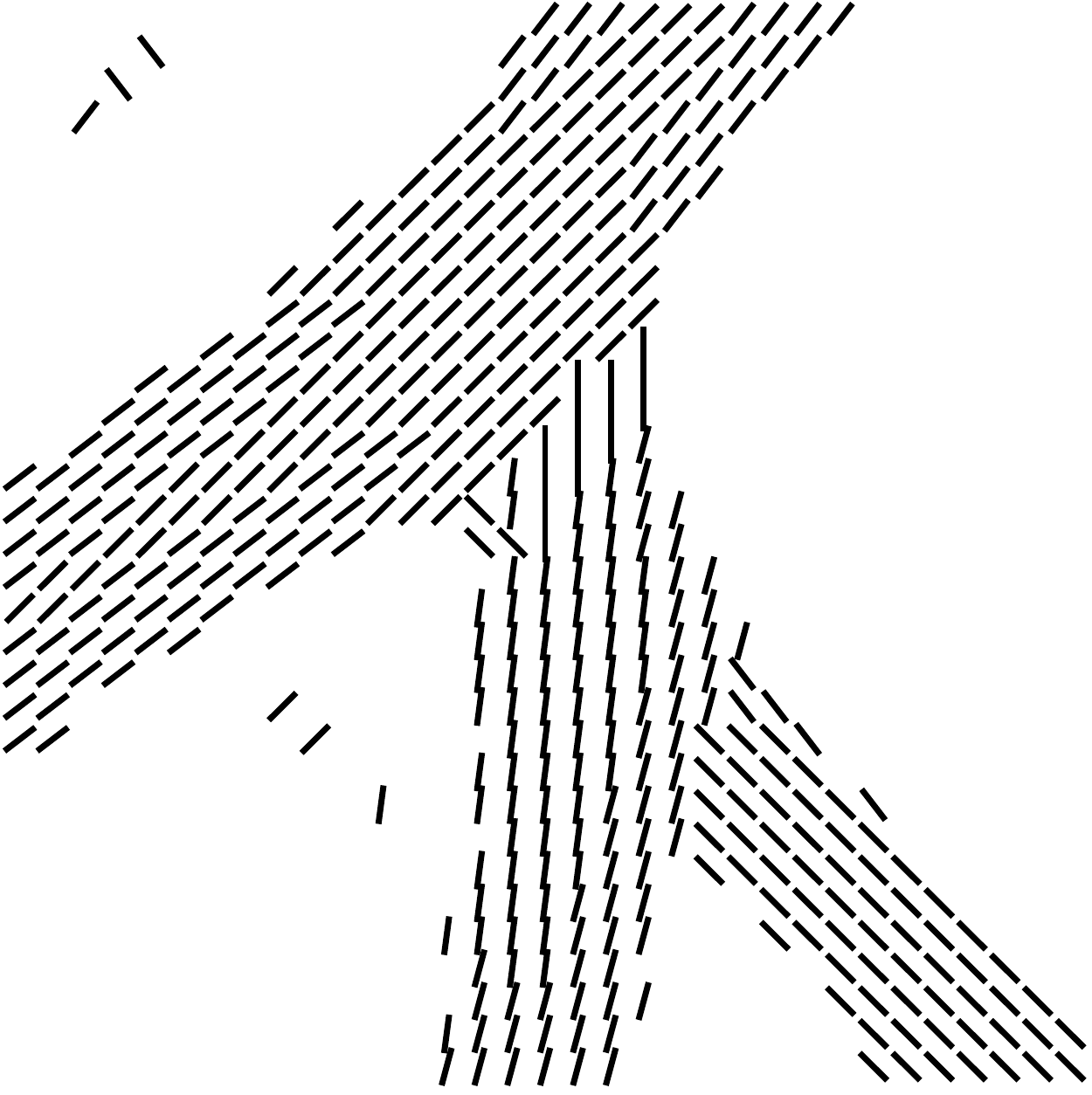}
		 \includegraphics[width=\textwidth]{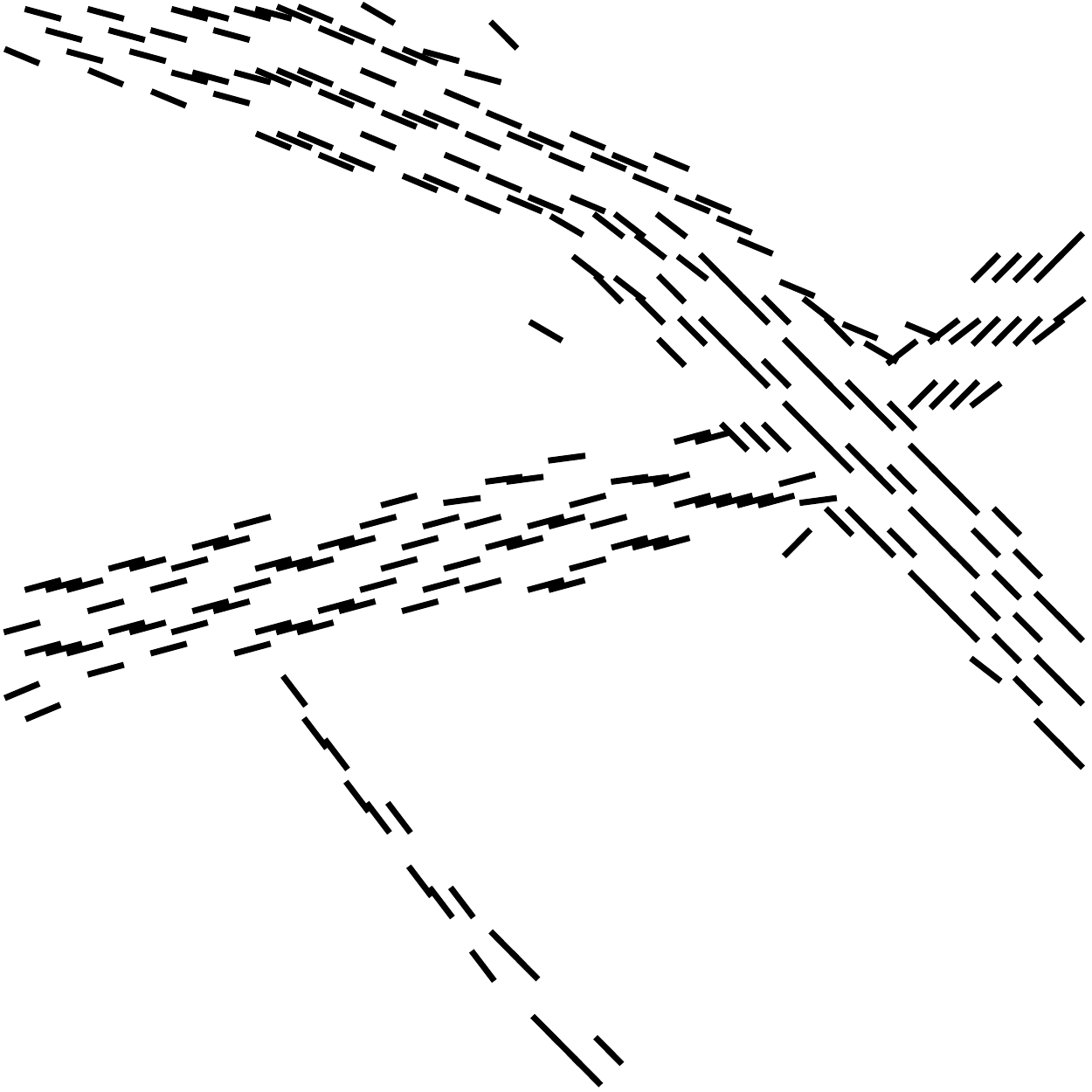}
		 \includegraphics[width=\textwidth]{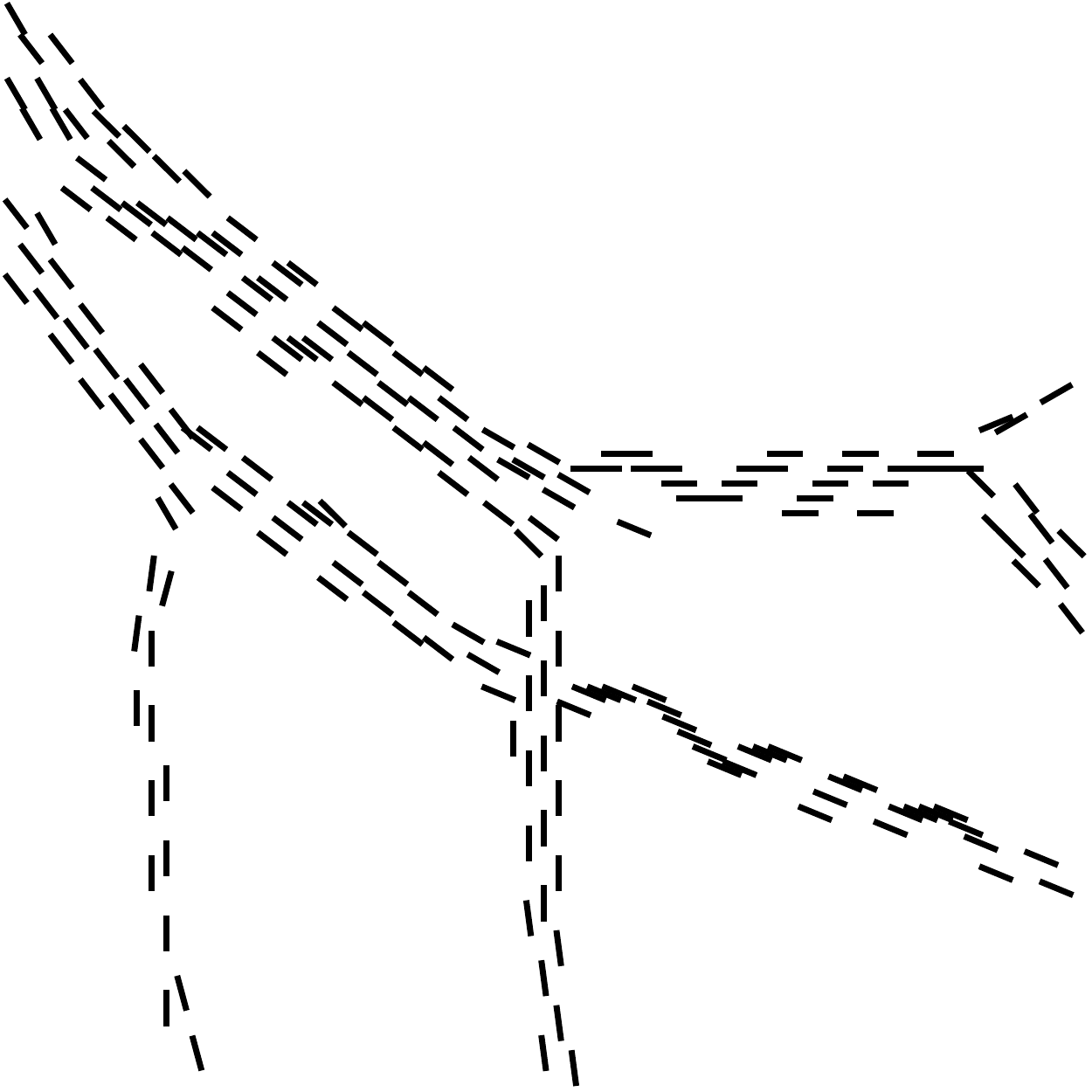}
		 \includegraphics[width=\textwidth]{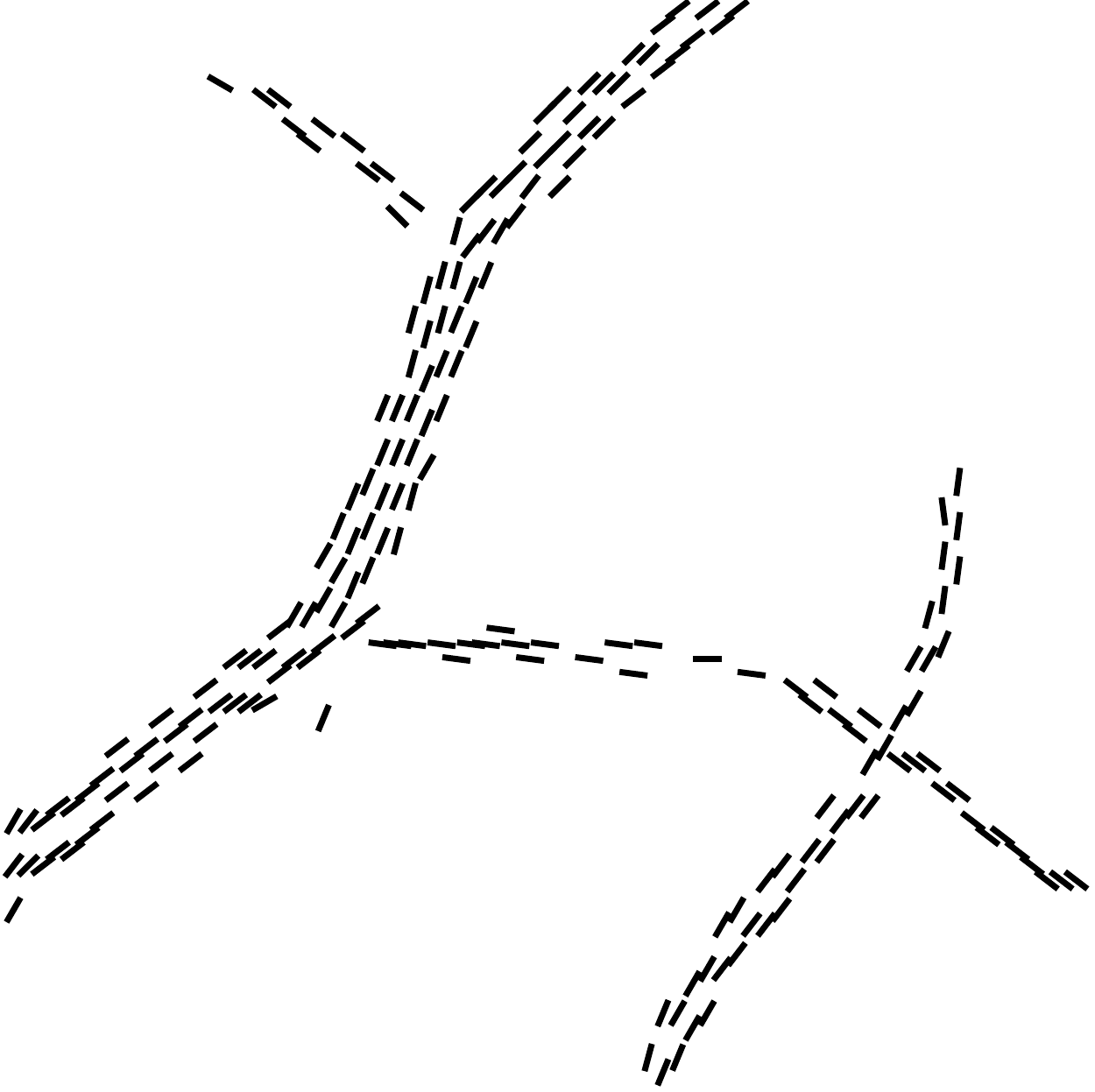}
		 \includegraphics[width=\textwidth]{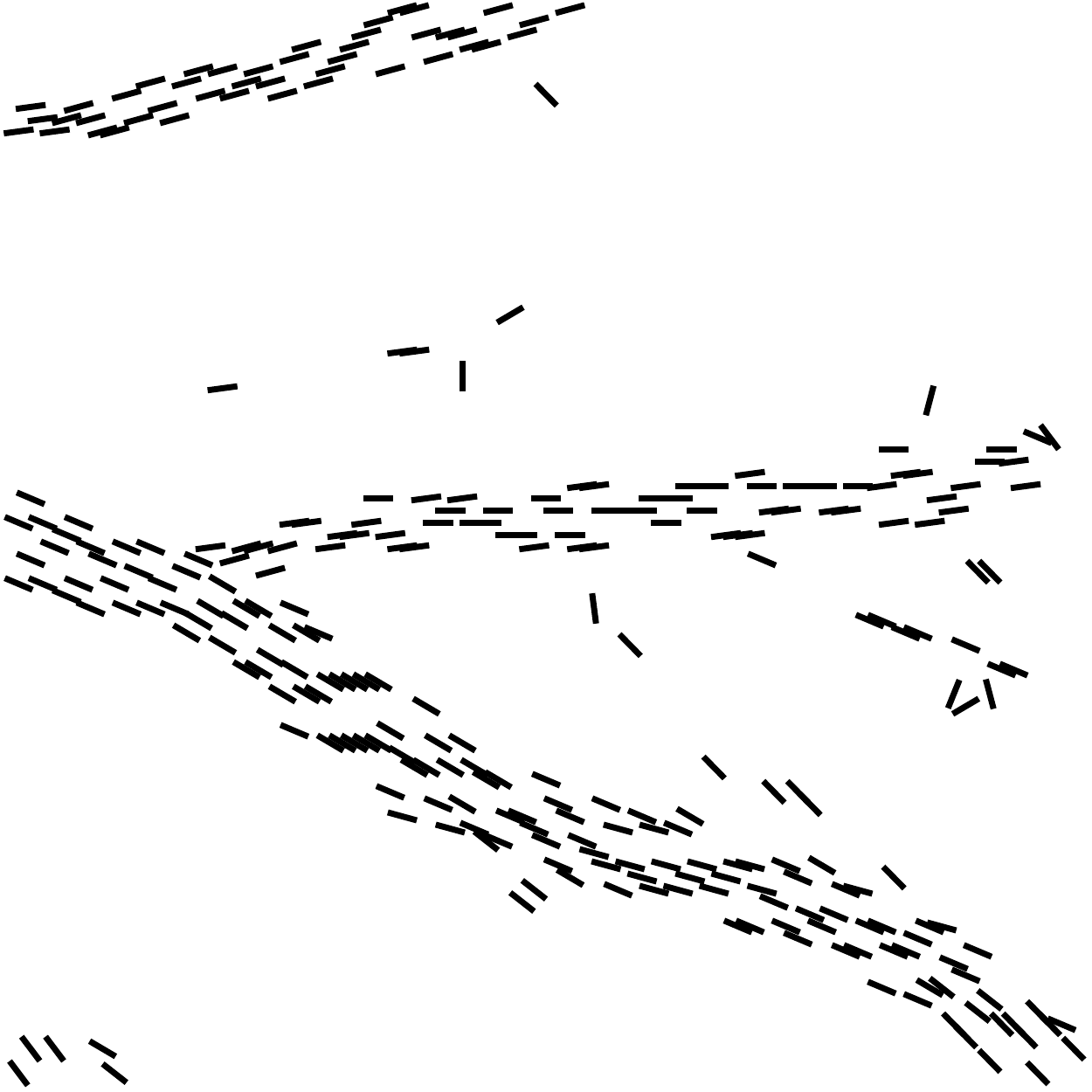}	 
		 \includegraphics[width=\textwidth]{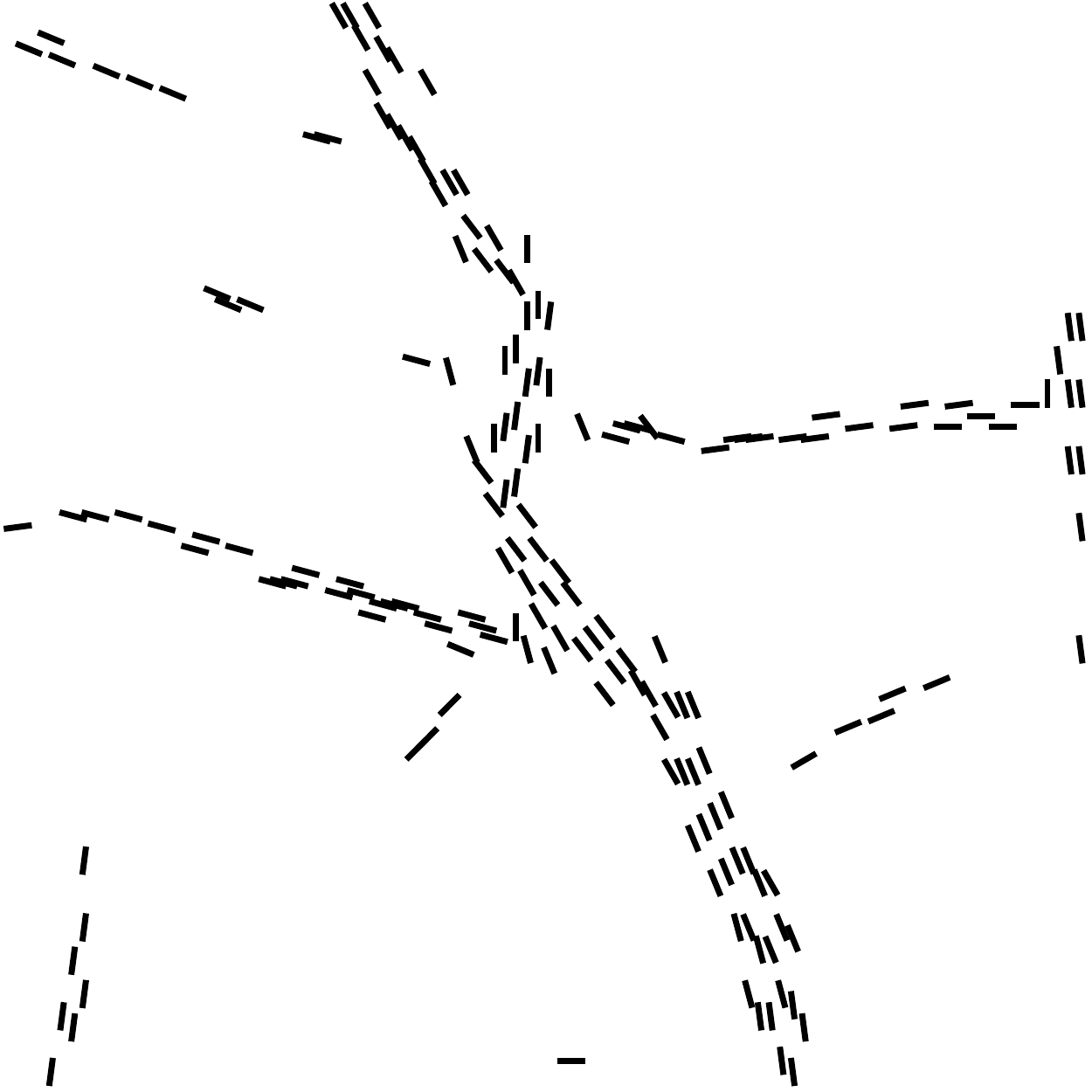}
		 \caption{}\label{fig:orientedSegments}
        \end{subfigure}    
          \begin{subfigure}[b]{0.9 in}
          \centering 
		 \includegraphics[width=\textwidth]{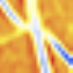}
		 \includegraphics[width=\textwidth]{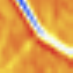}
		 \includegraphics[width=\textwidth]{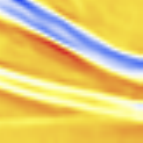}
		 \includegraphics[width=\textwidth]{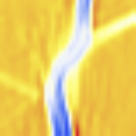}
        		 \includegraphics[width=\textwidth]{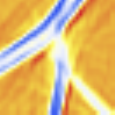} 
		 \includegraphics[width=\textwidth]{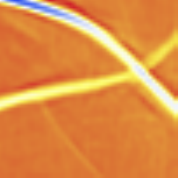}
		 \includegraphics[width=\textwidth]{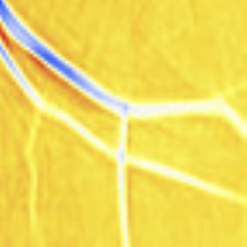}
		 \includegraphics[width=\textwidth]{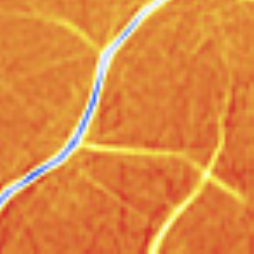}
		 \includegraphics[width=\textwidth]{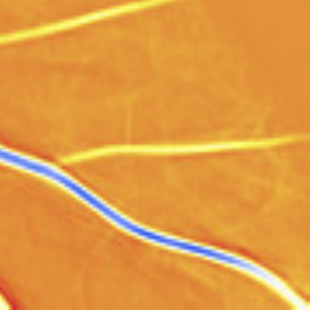}
		 \includegraphics[width=\textwidth]{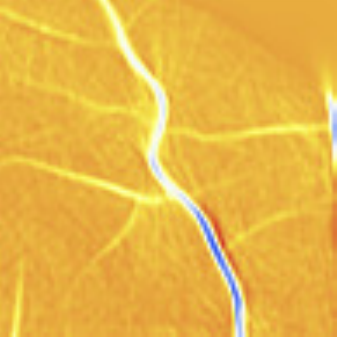}
		 \caption{}\label{fig:coloredintensity}
        \end{subfigure}   
        \begin{subfigure}[b]{0.9 in}
        \centering
		 \includegraphics[width=\textwidth]{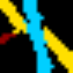}
		 \includegraphics[width=\textwidth]{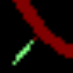}
		 \includegraphics[width=\textwidth]{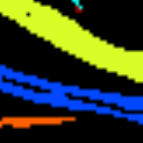}
		 \includegraphics[width=\textwidth]{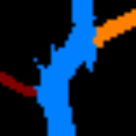}
        		 \includegraphics[width=\textwidth]{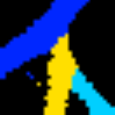}
		 \includegraphics[width=\textwidth]{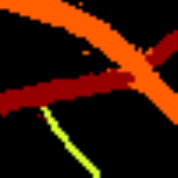}
		 \includegraphics[width=\textwidth]{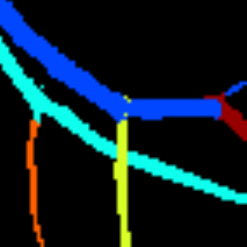}
		 \includegraphics[width=\textwidth]{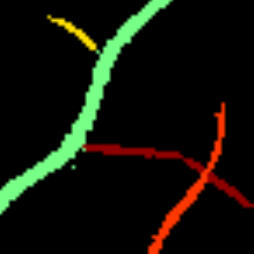}
		 \includegraphics[width=\textwidth]{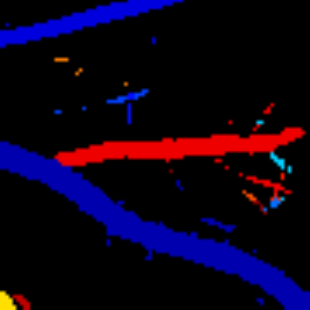}	 
		 \includegraphics[width=\textwidth]{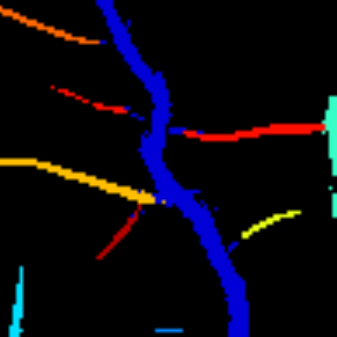}
		 \caption{ }\label{fig:results2}
        \end{subfigure}   
        \begin{subfigure}[b]{0.9 in}
        \centering
		  \includegraphics[width=\textwidth]{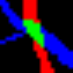}
		 \includegraphics[width=\textwidth]{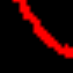}
		 \includegraphics[width=\textwidth]{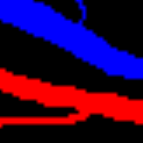}
		 \includegraphics[width=\textwidth]{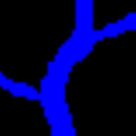}
        		 \includegraphics[width=\textwidth]{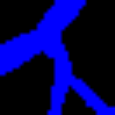}
		 \includegraphics[width=\textwidth]{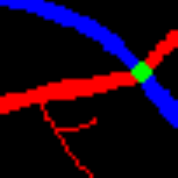}
		 \includegraphics[width=\textwidth]{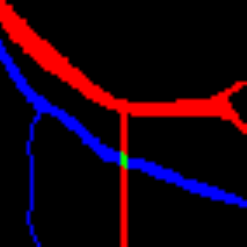}
		 \includegraphics[width=\textwidth]{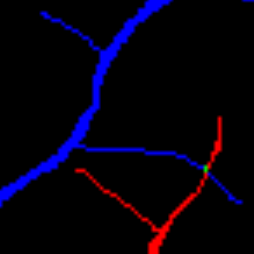}
		 \includegraphics[width=\textwidth]{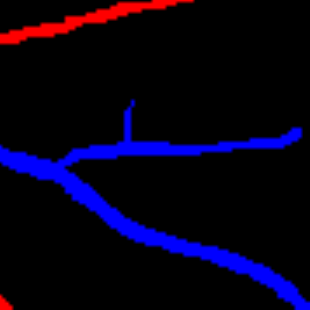}
		 \includegraphics[width=\textwidth]{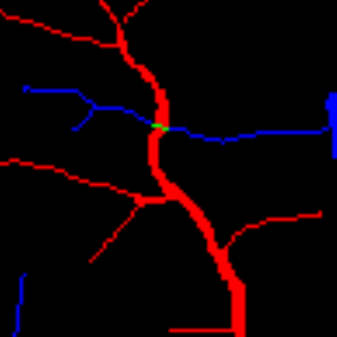}
		 \caption{ }\label{fig:avGT}
        \end{subfigure}    
        \caption{Sample image patches selected from the DRIVE dataset. Columns from left to right present the image patch at the green channel, segmented image, extracted orientation and intensity, clustering result and the artery/vein labels.}
        \label{fig:DifferentCases}
\end{figure*}

Some sample figures of these cases are depicted in Fig.~\ref{fig:DifferentCases}. For each example, the original gray scale enhanced image, hard  segmentation (locally thresholded), orientation and intensity information, and finally the clustering result together with artery/vein labels are depicted (Fig. \ref{fig:denoisedImg} to \ref{fig:avGT} respectively). Although the complexity of these patches is quite different in all cases, the salient groups are detected successfully. All the vessel pixels grouped as one unit have similarity in their orientations and intensities, and they follow the law of good continuation. Therefore, at each bifurcation or crossover point, two groups have been detected. 

In this figure, G1 is a good example of a crossing with a small angle. The method  not only differentiates well between vessels crossing each other even with a small crossing angle, but it also determines the order of vessels, being at the bottom or passing over in crossover regions.
The image patch in G2 is a good example showing the strength of the method in detecting small vessels. The detected small vessel in this image is even not annotated in the artery/vein ground truth. However, this detection is highly dependent on the soft segmented image and the threshold value used for obtaining the hard segmentation. If the small vessel is not detected in the soft segmentation or if a hight threshold value is selected, then it also will not be available in the final result. Other cases in this figure are good representations of the robustness of the method against the presence of a central vessel reflex (as in G3), interrupted lines (as in G10) or even noise (as in G9). In G9, noisy pixels are detected as individual units which are not similar to the other groups. They can be differentiated from others based on their sizes. If there are very few pixels in one group, then it can be considered as noise and removed. There are also several cases with complex junctions in this figure. Presence of multiple bifurcations in one image patch, or presence of several bifurcations close to the crossing points does not lead to wrong grouping results (as seen in G5, G6, G7, G8 and G10). 

The parameters used during the numerical simulations of the image patches shown in Fig.~\ref{fig:DifferentCases} and their corresponding sizes are presented in Table~\ref{tab:parameters}. 
For all experiments the values of $n$, $\epsilon$ and $\tau$ were set to $100000$, $0.1$ and $150$ respectively and they remained constant.
\begin{table}[htbp]
   \centering
   \caption{The parameters used in numerical simulation of the image patches shown in Fig.~\ref{fig:DifferentCases} and their corresponding sizes}
   \begin{tabular}{c*{4}{c}} 
    \hline
       Name 	&	size		&	$H$	& $\sigma$ 	& $\sigma_2$  \\
      \hline
      G1 			&$21\times 21$	&	7	&	0.02		&	0.3		 \\
      G2 			&$21\times 21$&	8	&	0.03		&	0.3		\\
      G3 			&$41\times 41$	&	10	&	0.03		&	0.1		 \\
      G4 			&$39\times 39$&	9	&	0.03		&	0.3		\\
      G5 			&$33\times 33$&	8	&	0.03		&	0.3		\\
      G6 			&$51\times 51$&	20	&	0.03		&	0.3		 \\
      G7 			&$71\times 71$&	17	&	0.07		&	0.3		\\
      G8 			&$73\times 73$&	24	&	0.03		&	0.3		\\
      G9 			&$89\times 89$&	30	&	0.03		&	0.3		 \\
      G10 			&$97\times 97$&	24	&	0.03		&	0.3		\\
       \hline
    \end{tabular}
   \label{tab:parameters}
\end{table}
The key parameters which are very effective in the final results are $H$, $\sigma$ and $\sigma_2$. $H$ and $\sigma$ determine the shape of the kernel. Based on the experiments, the appropriate value for the number of steps of the random path generation is approximately $1/3$ of the image width. Selecting this parameter correctly is very important in connecting the interrupted lines. The parameters  $\sigma$ and $\sigma_2$ which determine the propagation variance in the $\theta$ direction and the effect of the intensity-based similarity term do not have a large sensitivity to variation. To quantify this, the mean and variance of these two parameters for each of the above-mentioned categories are calculated and presented in Table~\ref{tbl:avgParam}. Since the selected patches have varying sizes and $H$ is dependent on that, this parameter is not presented in this table. Moreover, to evaluate the performance of the method, we introduced the correct detection rate ($CDR$) as the percentage of correctly grouped image patches for each category. These values are presented in Table \ref{tbl:avgParam}. By considering higher number of image patches per category the $CDR$ values will be more realistic.
\begin{table}[htbp]
   \centering
   \caption{The correct detection rate and the mean and variance of $\sigma$ and $\sigma_2$ used in numerical simulation for each category}
   \begin{tabular}{c*{5}{c}} 
   	 \hline
      \multirow{2}{*}{Category} & \multirow{2}{*}{CDR\%} &\multicolumn{2}{c}{$\sigma$} &	\multicolumn{2}{c}{$\sigma_2$} 	\\
      						&				    &			mean 	&	variance	& 	mean		&	variance		\\
      \hline
      A 				&			 85   	   &	0.032	&	0.0001	&		0.28		&		0.0039	\\
      B 				&			95      &	0.033	&$\simeq$ 0	&		0.3		&	$\simeq$ 0	\\
      C 				&			85		    &	0.0269	&$\simeq$ 0	&		0.22		&		0.01		\\
      D 		 		&			75		    &	0.035	&	0.00013	&		0.248	&		0.0125	\\
      E 				&			95		    &	0.03		&$\simeq$ 0	&		0.3		&	$\simeq$ 0	\\
       \hline
    \end{tabular}
   \label{tbl:avgParam}
\end{table}

\begin{figure*}[htbp]
\centering
	\begin{subfigure}[b]{0.9 in}
	 \includegraphics[width= \columnwidth]{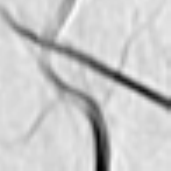}\caption{}\label{fig:hcsampledenoised}
	 \end{subfigure}
	 \begin{subfigure}[b]{0.9 in}
	 \includegraphics[width=\columnwidth]{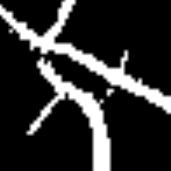}\caption{}\label{fig:hcsamplehardSeg}
	 \end{subfigure}
	  \begin{subfigure}[b]{0.9 in}
	 \includegraphics[width=\columnwidth]{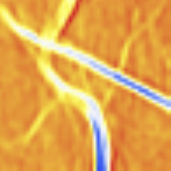}\caption{}\label{fig:hcsampleIntensity}
	 \end{subfigure}
	 \begin{subfigure}[b]{0.9 in}
	 \includegraphics[width=\columnwidth]{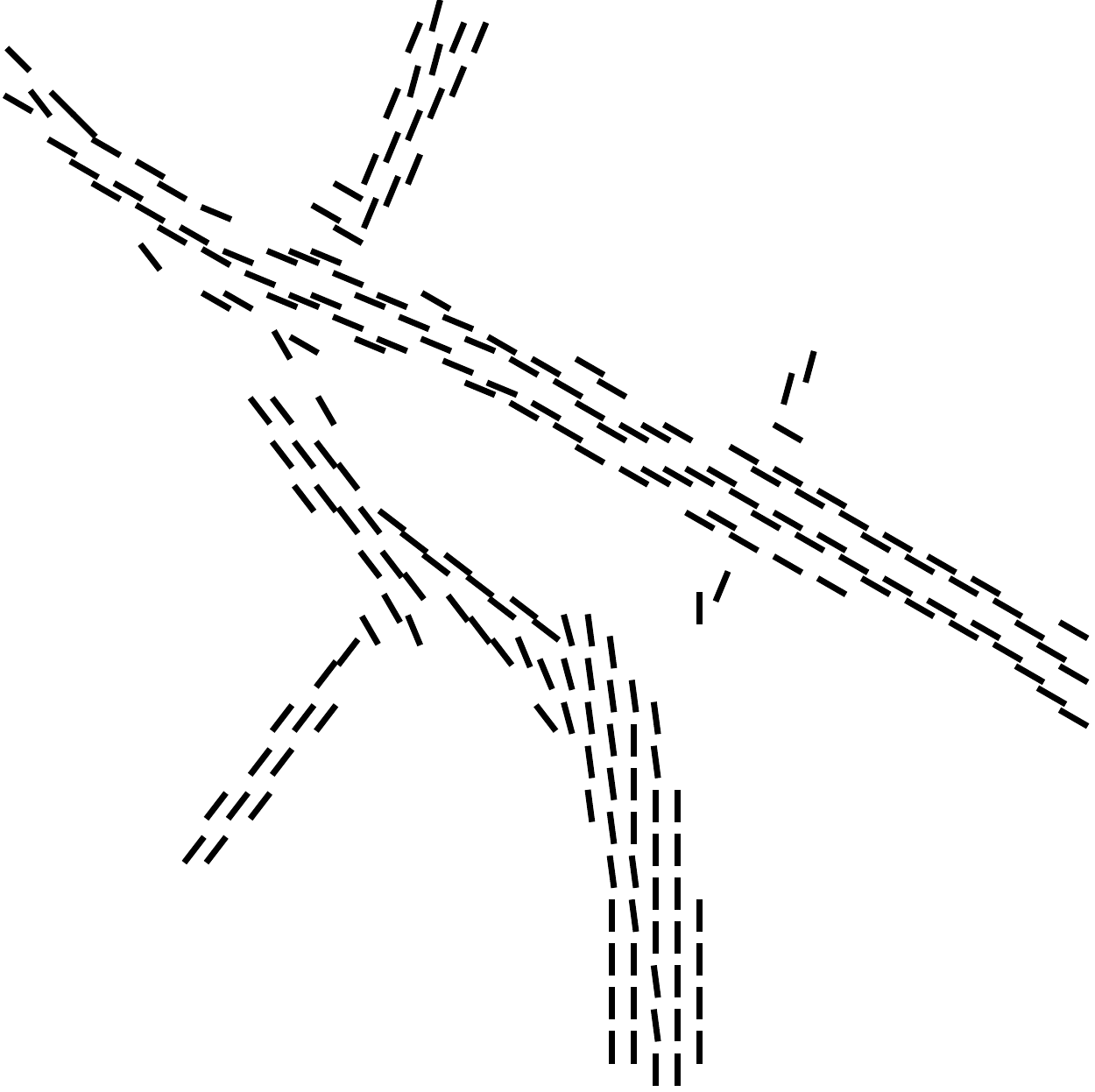}\caption{}\label{fig:hcsampleOrientation}
	 \end{subfigure}
	  \begin{subfigure}[b]{0.9 in}
	 \includegraphics[width=\columnwidth]{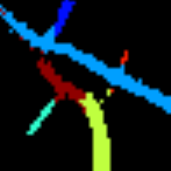}\caption{}\label{fig:hcsampleResult2}
	 \end{subfigure}
	 \begin{subfigure}[b]{0.9 in}
	 \includegraphics[width=\columnwidth]{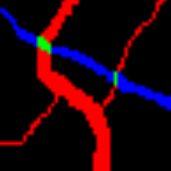}\caption{}\label{fig:hcsampleavGT}
	 \end{subfigure}
	 \caption{Failure of clustering in presence of highly curved vessels. Columns from left to right: enhanced image; its segmentation; orientation and intensity information; clustering result and the artery/vein labels.}
	 \label{fig:highcurvature}
\end{figure*}

If there are some high curvature vessels, then depending on their curvature increasing $\sigma$ might help in preserving the continuity of the vessel. As an example, G4 in Fig.~\ref{fig:DifferentCases} is relatively more curved compared to the other cases, but the clustering works perfectly in this case. However, for some cases it does not solve the problem totally, and other kernels need to be considered for preserving the continuity. An example $49\times 49$ image patch with a highly curved vessel is shown in Fig~\ref{fig:highcurvature}, where the method fails in clustering the vessels correctly. The parameters used for this case are $H=16$, $\sigma=0.03$ and $\sigma_2 = 0.3$ 

\begin{figure*}[htbp]
\centering
	\begin{subfigure}[b]{0.8 in}
	 \includegraphics[width= \columnwidth]{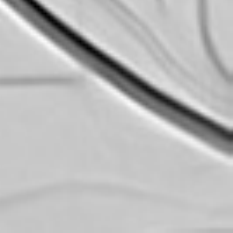}\caption{}\label{fig:IEsampledenoised}
	 \end{subfigure}
	 \begin{subfigure}[b]{0.8 in}
	 \includegraphics[width=\columnwidth]{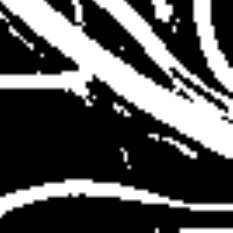}\caption{}\label{fig:IEsamplehardSeg}
	 \end{subfigure}
	  \begin{subfigure}[b]{0.8 in}
	 \includegraphics[width=\columnwidth]{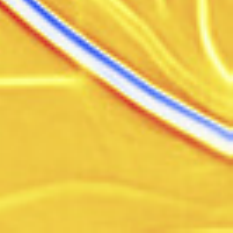}\caption{}\label{fig:IEsampleIntensity}
	 \end{subfigure}
	 \begin{subfigure}[b]{0.8 in}
	 \includegraphics[width=\columnwidth]{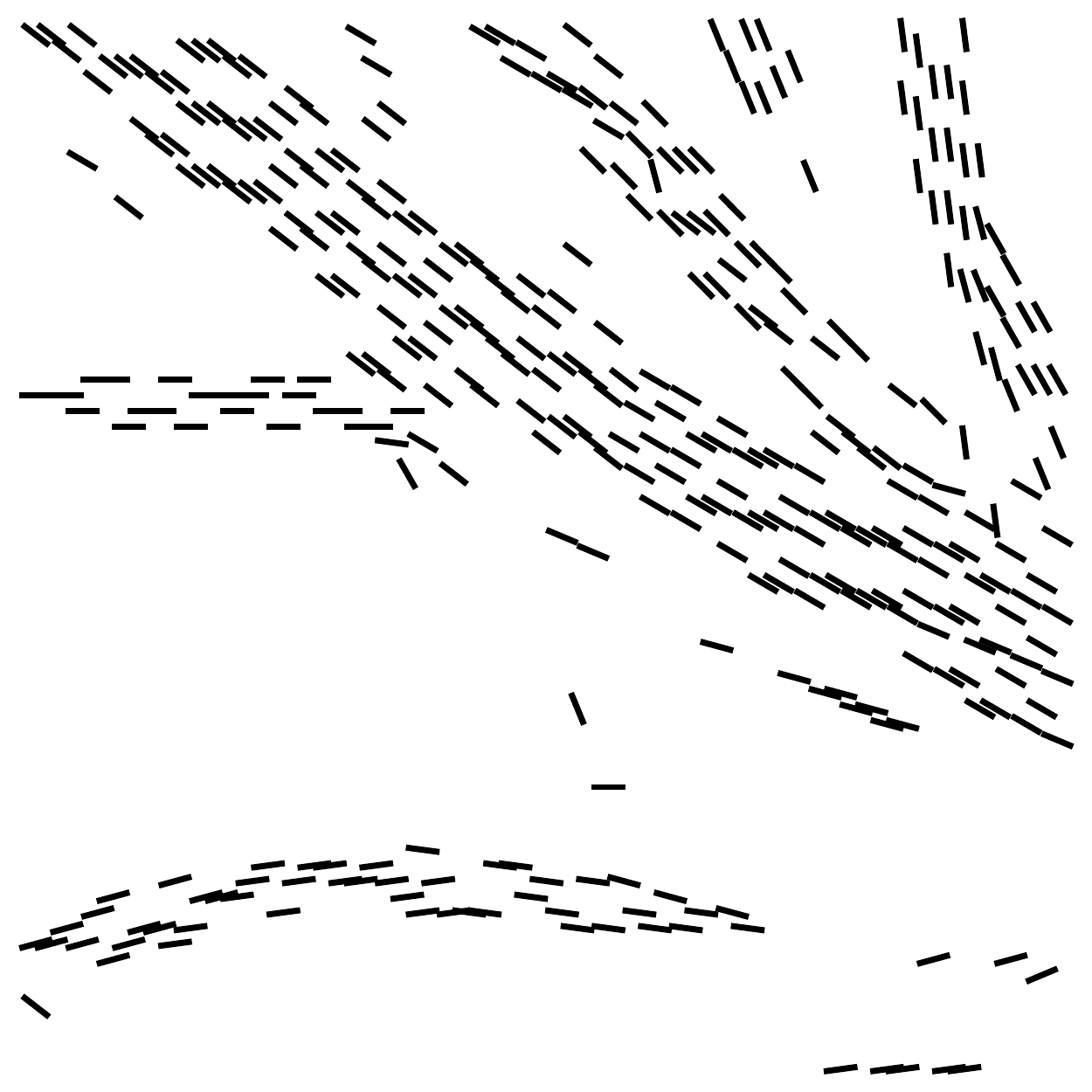}\caption{}\label{fig:IEsampleOrientation}
	 \end{subfigure}
	  \begin{subfigure}[b]{0.8 in}
	 \includegraphics[width=\columnwidth]{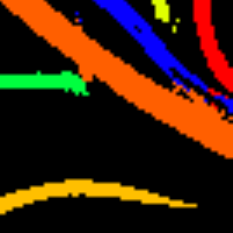}\caption{}\label{fig:IEsampleResult2}
	 \end{subfigure}
	 \begin{subfigure}[b]{0.8 in}
	 \includegraphics[width=\columnwidth]{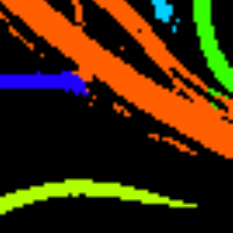}\caption{}\label{fig:IEsampleWrongResult}
	 \end{subfigure}
	 \begin{subfigure}[b]{0.8 in}
	 \includegraphics[width=\columnwidth]{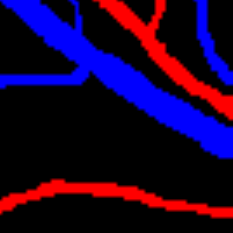}\caption{}\label{fig:IEsampleavGT}
	 \end{subfigure}
	 \caption{The effect of including intensity term in calculating the connectivity kernel. 
	 Columns from left to right: enhanced image; its segmentation; orientation and intensity information; correct and wrong clustering results; and the artery/vein labels. $\sigma_2=0.3$ and $1$ are used for (\subref{fig:IEsampleResult2}) and (\subref{fig:IEsampleWrongResult}) respectively}
	 \label{fig:IntensityEffect}
\end{figure*}

Even though the intensities of arteries and veins in the gray scale enhanced image are very close to each other in some images, adding the intensity term in calculating the final affinity matrix is crucial. By decreasing the value of $\sigma_2$, the distance between intensities gets a higher value and it helps in differentiating better between the groups. Fig.~\ref{fig:IntensityEffect} represents a sample $67\times 67$ image patch, which includes two nearby parallel vessels with similar orientations. Fig~\ref{fig:IEsampleResult2} and \ref{fig:IEsampleWrongResult} show the correct and wrong clustering results obtained by changing $\sigma_2$ from $0.3$ to $1$. All other parameters have not changed ($H=24$, $\sigma=0.02$ and $n=100000$). The other important difference between these two results is that the noisy pixels close to the thicker vessel have been totally removed in the correct result. Although they seem to be oriented with the thick vessel their intensities are totally different. Therefore, by increasing the effect of intensity, they are clustered as several small groups and removed in the final step of the spectral clustering algorithm. 

\section{Conclusion and Future Work}
\label{sec:conclusion}

In this work, we have presented a novel semi-automatic technique inspired by the geometry of the primary visual cortex to find and group different perceptual units in retinal images using spectral methods.
Computing eigenvectors of affinity matrices, which are formed using the connectivity kernel, leads to the final grouping. The connectivity kernel represents the connectivity between all lifted points to the 4-dimensional feature space of positions, orientations and intensities, and it presents a good model for the Gestalt law of good continuation. Thus, the perceptual units in retinal images are the individual blood vessels having low variation in their orientations and intensities. 
%

The proposed method allows finding accurate junction positions, which is the position where two groups meet or cross each other. The main application of these connectivity analyses would be in modelling the retinal vasculature as a set of tree networks. The main graph constructed by these trees would be very informative in analyzing the topological behaviour of retinal vasculature which is useful in diagnosis and prognosis of several diseases especially in automated application in large scale screening programs. 

The detection of small vessels highly depends on the quality of the soft segmentation, not the hard segmentation. These vessels could easily be differentiated from noise based on the size of the group. Noisy pixels have random orientations and intensities and they build smaller groups. 
Our method represents some limitations at blood vessels with high curvature. One possible solution is to merge the two detected perceptual units and form one unique unit, if there are no junctions at these locations. The other stronger extension is to use other kernels that take into account the curvature of structures in addition to positions and orientations. Moreover, it is also possible to enrich the affinity matrix with other terms e.g. the principle curvature of the multiscale Hessian (ridgeness or vesselness similarity). All these solutions will be investigated in the future. 

With this model we have analyzed many challenging cases, such as bifurcations, crossovers, small and disconnected vessels in retinal vessel segmentations. These cases not only have been reported to create tracing errors in the state-of-the-art techniques, but also are very informative for the clinical studies. Based on the results shown in the numerical simulations, the method is successful in detecting the salient groups in retinal images, and robust against noise, central vessel reflex, interruptions in vessel segments, presence of multiple junctions in a small area and presence of nearby parallel vessels. For this reason, this can be considered as an excellent quantitative model for the constitution of perceptual units in retinal images. To the best of our knowledge, this is the first time that the vessel connectivities in such complex situations are solved by one single solution perfectly.

\begin{acknowledgements}
This project has received funding from the European Union's Seventh Framework Programme, Marie Curie Actions- Initial Training Network, under grant agreement $n^o 607643$, ``Metric Analysis For Emergent Technologies (MAnET)". It was also supported by the H\'e Programme of Innovation, which is partly financed by the Netherlands Organization for Scientific research (NWO) under grant $n^o 629.001.003$. 
\end{acknowledgements}

\bibliographystyle{spmpsci}      
\bibliography{manuscript}   

\begin{thebibliography}{10}
\providecommand{\url}[1]{{#1}}
\providecommand{\urlprefix}{URL }
\expandafter\ifx\csname urlstyle\endcsname\relax
  \providecommand{\doi}[1]{DOI~\discretionary{}{}{}#1}\else
  \providecommand{\doi}{DOI~\discretionary{}{}{}\begingroup
  \urlstyle{rm}\Url}\fi

\bibitem{abbasi2015biologically}
Abbasi-Sureshjani, S., Smit-Ockeloen, I., Zhang, J., ter Haar~Romeny, B.:
  Biologically-inspired supervised vasculature segmentation in {SLO} retinal
  fundus images.
\newblock In: Image Analysis and Recognition, vol. 9164, pp. 325--334. Springer
  (2015)

\bibitem{al2009active}
Al-Diri, B., Hunter, A., Steel, D.: An active contour model for segmenting and
  measuring retinal vessels.
\newblock {IEEE} T. Med. Imaging \textbf{28}(9), 1488--1497 (2009)

\bibitem{august2000curve}
August, J., Zucker, S.W.: The curve indicator random field: Curve organization
  via edge correlation.
\newblock In: Perceptual organization for artificial vision systems, pp.
  265--288. Springer (2000)

\bibitem{august2003sketches}
August, J., Zucker, S.W.: Sketches with curvature: The curve indicator random
  field and {M}arkov processes.
\newblock {IEEE} T. Pattern. Anal. \textbf{25}(4), 387--400 (2003)

\bibitem{Bekkers:2014aa}
Bekkers, E., Duits, R., Berendschot, T., ter Haar~Romeny, B.: A
  multi-orientation analysis approach to retinal vessel tracking.
\newblock J. Math. Imaging Vis. \textbf{49}(3), 583--610 (2014)

\bibitem{boscain2012anthropomorphic}
Boscain, U., Duplaix, J., Gauthier, J.P., Rossi, F.: Anthropomorphic image
  reconstruction via hypoelliptic diffusion.
\newblock SIAM J. Control Optim. \textbf{50}(3), 1309--1336 (2012)

\bibitem{bosking1997orientation}
Bosking, W.H., Zhang, Y., Schofield, B., Fitzpatrick, D.: Orientation
  selectivity and the arrangement of horizontal connections in tree shrew
  striate cortex.
\newblock J. Neurosci. \textbf{17}(6), 2112--2127 (1997)

\bibitem{bressloff2002geometric}
Bressloff, P.C., Cowan, J.D., Golubitsky, M., Thomas, P.J., Wiener, M.C.: What
  geometric visual hallucinations tell us about the visual cortex.
\newblock Neural. Comput. \textbf{14}(3), 473--491 (2002)

\bibitem{buhler2004geometric}
B{\"u}hler, K., Felkel, P., La~Cruz, A.: Geometric methods for vessel
  visualization and quantification---a survey.
\newblock In: Geometric Modeling for Scientific Visualization, Math. Visual.,
  pp. 399--419. Springer Berlin Heidelberg (2004)

\bibitem{can1999rapid}
Can, A., Shen, H., Turner, J.N., Tanenbaum, H.L., Roysam, B.: Rapid automated
  tracing and feature extraction from retinal fundus images using direct
  exploratory algorithms.
\newblock {IEEE} T. Inf. Technol. B. \textbf{3}(2), 125--138 (1999)

\bibitem{chapman2002peripheral}
Chapman, N., Dell'Omo, G., Sartini, M., Witt, N., Hughes, A., Thom, S.,
  Pedrinelli, R.: Peripheral vascular disease is associated with abnormal
  arteriolar diameter relationships at bifurcations in the human retina.
\newblock Clin. Sci. \textbf{103}(2), 111--116 (2002)

\bibitem{chutatape1998retinal}
Chutatape, O., Zheng, L., Krishnan, S.: Retinal blood vessel detection and
  tracking by matched {G}aussian and {K}alman filters.
\newblock In: Engineering in Medicine and Biology Society, 1998. Proceedings of
  the 20th Annual International Conference of the IEEE, vol.~6, pp. 3144--3149.
  IEEE (1998)

\bibitem{citti2006cortical}
Citti, G., Sarti, A.: A cortical based model of perceptual completion in the
  roto-translation space.
\newblock J. Math. Imaging Vis. \textbf{24}(3), 307--326 (2006)

\bibitem{cocci2015cortical}
Cocci, G., Barbieri, D., Citti, G., Sarti, A.: Cortical spatio-temporal
  dimensionality reduction for visual grouping.
\newblock Neural. Comput. \textbf{27}(6), 1252--1293 (2015)

\bibitem{coifman2006diffusion}
Coifman, R.R., Lafon, S.: Diffusion maps.
\newblock Appl. Comput. Harmon. A. \textbf{21}(1), 5--30 (2006)

\bibitem{de2014tracing}
De, J., Li, H., Cheng, L.: Tracing retinal vessel trees by transductive
  inference.
\newblock {BMC} Bioinformatics \textbf{15}(1), 20 (2014)

\bibitem{de2013automated}
De, J., Ma, T., Li, H., Dash, M., Li, C.: Automated Tracing of Retinal Blood
  Vessels Using Graphical Models, \emph{LNCS}, vol. 7944, pp. 277--289.
\newblock Springer Berlin Heidelberg (2013)

\bibitem{delibasis2010automatic}
Delibasis, K.K., Kechriniotis, A.I., Tsonos, C., Assimakis, N.: Automatic
  model-based tracing algorithm for vessel segmentation and diameter
  estimation.
\newblock Comput. Meth. Prog. Bio. \textbf{100}(2), 108--122 (2010)

\bibitem{DuitsCASA2005}
Duits, R., van Almsick, M.: The explicit solutions of linear left-invariant
  second order stochastic evolution equations on the 2{D}-{E}uclidean motion
  group.
\newblock Technical Report CASA-report~43, Eindhoven University of Technology
  Dep. of mathematics and computer science (2005)

\bibitem{duits2008explicit}
Duits, R., van Almsick, M.: The explicit solutions of linear left-invariant
  second order stochastic evolution equations on the {2D} {E}uclidean motion
  group.
\newblock Q. Appl. Math. \textbf{66}(1), 27--68 (2008)

\bibitem{duits2007invertible}
Duits, R., Duits, M., van Almsick, M., ter Haar~Romeny, B.: Invertible
  orientation scores as an application of generalized wavelet theory.
\newblock S. Mach. Perc. \textbf{17}(1), 42--75 (2007)

\bibitem{duits2007image}
Duits, R., Felsberg, M., Granlund, G., ter Haar~Romeny, B.: Image analysis and
  reconstruction using a wavelet transform constructed from a reducible
  representation of the {E}uclidean motion group.
\newblock Int. J. Comput. Vision \textbf{72}(1), 79--102 (2007)

\bibitem{duits2010left}
Duits, R., Franken, E.: Left-invariant parabolic evolutions on {SE(2)} and
  contour enhancement via invertible orientation scores---part i: Linear
  left-invariant diffusion equations on {SE(2)}.
\newblock Q. Appl. Math. \textbf{68}(2), 255--292 (2010)

\bibitem{duits2013evolution}
Duits, R., F{\"u}hr, H., Janssen, B., Bruurmijn, M., Florack, L., van Assen,
  H.: Evolution equations on gabor transforms and their applications.
\newblock Appl. Comput. Harmon. A. \textbf{35}(3), 483--526 (2013)

\bibitem{faugeras2009persistent}
Faugeras, O., Veltz, R., Grimbert, F.: Persistent neural states: stationary
  localized activity patterns in nonlinear continuous n-population,
  q-dimensional neural networks.
\newblock Neural. Comput. \textbf{21}(1), 147--187 (2009)

\bibitem{felkel2001vessel}
Felkel, P., Wegenkittl, R., Kanitsar, A.: Vessel tracking in peripheral {CTA}
  datasets-an overview.
\newblock In: Computer Graphics, Spring Conference on, 2001., pp. 232--239.
  IEEE (2001)

\bibitem{field1993contour}
Field, D.J., Hayes, A., Hess, R.F.: Contour integration by the human visual
  system: Evidence for a local ``association field''.
\newblock Vision Res. \textbf{33}(2), 173--193 (1993)

\bibitem{foracchia2001extraction}
Foracchia, M., Grisan, E., Ruggeri, A.: Extraction and quantitative description
  of vessel features in hypertensive retinopathy fundus images.
\newblock In: Book Abstracts 2nd International Workshop on Computer Assisted
  Fundus Image Analysis, p.~6 (2001)

\bibitem{Foracchia2005}
Foracchia, M., Grisan, E., Ruggeri, A.: Luminosity and contrast normalization
  in retinal images.
\newblock Med. Image. Anal. \textbf{9}(3), 179 -- 190 (2005)

\bibitem{franken2009crossing}
Franken, E., Duits, R.: Crossing-preserving coherence-enhancing diffusion on
  invertible orientation scores.
\newblock Int. J. Comput. Vision \textbf{85}(3), 253--278 (2009)

\bibitem{Fraz2012407}
Fraz, M., Remagnino, P., Hoppe, A., Uyyanonvara, B., Rudnicka, A., Owen, C.,
  Barman, S.: Blood vessel segmentation methodologies in retinal images -- a
  survey.
\newblock Comput. Meth. Prog. Bio. \textbf{108}(1), 407 -- 433 (2012)

\bibitem{fruttiger2007development}
Fruttiger, M.: Development of the retinal vasculature.
\newblock Angiogenesis \textbf{10}(2), 77--88 (2007)

\bibitem{fuhr2005abstract}
F{\"u}hr, H.: Abstract harmonic analysis of continuous wavelet transforms.
\newblock Springer (2005)

\bibitem{gonzalez2010delineating}
Gonz{\'a}lez, G., T{\"u}retken, E., Fleuret, F., Fua, P.: Delineating trees in
  noisy {2D} images and {3D} image-stacks.
\newblock In: IEEE Conference on Computer Vision and Pattern Recognition
  (CVPR), pp. 2799--2806 (2010)

\bibitem{habib2014association}
Habib, M.S., Al-Diri, B., Hunter, A., Steel, D.H.: The association between
  retinal vascular geometry changes and diabetic retinopathy and their role in
  prediction of progression-an exploratory study.
\newblock {BMC} Ophthalmol. \textbf{14}(1), 89 (2014)

\bibitem{higham2001algorithmic}
Higham, D.J.: An algorithmic introduction to numerical simulation of stochastic
  differential equations.
\newblock {SIAM} Rev. \textbf{43}(3), 525--546 (2001)

\bibitem{hoffman1989visual}
Hoffman, W.C.: The visual cortex is a contact bundle.
\newblock Appl. Math. Comput. \textbf{32}(2), 137--167 (1989)

\bibitem{hu2011globalization}
Hu, F.B.: Globalization of diabetes the role of diet, lifestyle, and genes.
\newblock Diabetes Care \textbf{34}(6), 1249--1257 (2011)

\bibitem{hu2013automated}
Hu, Q., Abr{\`a}moff, M.D., Garvin, M.K.: Automated Separation of Binary
  Overlapping Trees in Low-Contrast Color Retinal Images, \emph{LNCS}, vol.
  8150, pp. 436--443.
\newblock Springer Berlin Heidelberg (2013)

\bibitem{hubbard1999methods}
Hubbard, L.D., Brothers, R.J., King, W.N., Clegg, L.X., Klein, R., Cooper,
  L.S., Sharrett, A.R., Davis, M.D., Cai, J., in~Communities Study~Group, A.R.,
  et~al.: Methods for evaluation of retinal microvascular abnormalities
  associated with hypertension/sclerosis in the atherosclerosis risk in
  communities study.
\newblock Ophthalmology \textbf{106}(12), 2269--2280 (1999)

\bibitem{hubel1962receptive}
Hubel, D.H., Wiesel, T.N.: Receptive fields, binocular interaction and
  functional architecture in the cat's visual cortex.
\newblock J. Physiol. \textbf{160}(1), 106 (1962)

\bibitem{hubel1977ferrier}
Hubel, D.H., Wiesel, T.N.: Ferrier lecture: Functional architecture of macaque
  monkey visual cortex.
\newblock P. Roy. Soc. Lond. B Bio. \textbf{198}(1130), 1--59 (1977)

\bibitem{ICOGuidelines}
The International Council of Ophthalmology: ICO Guidelines for Diabetic Eye
  Care (2014)

\bibitem{joshi2011automated}
Joshi, V.S., Garvin, M.K., Reinhardt, J.M., Abramoff, M.D.: Automated method
  for the identification and analysis of vascular tree structures in retinal
  vessel network.
\newblock In: SPIE Medical Imaging, pp. 79,630I--79,630I. International Society
  for Optics and Photonics (2011)

\bibitem{koenderink1987representation}
Koenderink, J.J., van Doorn, A.J.: Representation of local geometry in the
  visual system.
\newblock Biol. Cybern. \textbf{55}(6), 367--375 (1987)

\bibitem{lowell2004measurement}
Lowell, J., Hunter, A., Steel, D., Basu, A., Ryder, R., Kennedy, R.L.:
  Measurement of retinal vessel widths from fundus images based on 2-{D}
  modeling.
\newblock {IEEE} T. Med. Imaging \textbf{23}(10), 1196--1204 (2004)

\bibitem{meila2001random}
Meila, M., Shi, J.: A random walks view of spectral segmentation.
\newblock In: Artif. Int. Stat. (AISTATS) (2001)

\bibitem{mozaffarian2015heart}
Mozaffarian, D., Benjamin, E.J., Go, A.S., Arnett, D.K., Blaha, M.J., Cushman,
  M., de~Ferranti, S., Despres, J.P., Fullerton, H.J., Howard, V.J., et~al.:
  Heart disease and stroke statistics-2015 update: a report from the {A}merican
  {H}eart {A}ssociation.
\newblock Circulation \textbf{131}(4), e29 (2015)

\bibitem{mumford1994elastica}
Mumford, D.: Elastica and computer vision, pp. 491--506.
\newblock Springer New York (1994)

\bibitem{narasimha2008improved}
Narasimha-Iyer, H., Mahadevan, V., Beach, J.M., Roysam, B.: Improved detection
  of the central reflex in retinal vessels using a generalized dual-{G}aussian
  model and robust hypothesis testing.
\newblock {IEEE} T. Inf. Technol. B. \textbf{12}(3), 406--410 (2008)

\bibitem{ng2002spectral}
Ng, A.Y., Jordan, M.I., Weiss, Y., et~al.: On spectral clustering: Analysis and
  an algorithm.
\newblock Adv. Neur. In. \textbf{2}, 849--856 (2002)

\bibitem{olsen2011convolution}
Olsen, M.A., Hartung, D., Busch, C., Larsen, R.: Convolution approach for
  feature detection in topological skeletons obtained from vascular patterns.
\newblock In: {IEEE} Workshop on Computational Intelligence in Biometrics and
  Identity Management ({CIBIM}), pp. 163--167 (2011)

\bibitem{otsu1975threshold}
Otsu, N.: A threshold selection method from gray-level histograms.
\newblock {IEEE} T. Syst. Man. Cyb. \textbf{9}(1), 62--66 (1979)

\bibitem{parent1989trace}
Parent, P., Zucker, S.W.: Trace inference, curvature consistency, and curve
  detection.
\newblock {IEEE} T. Pattern. Anal. \textbf{11}(8), 823--839 (1989)

\bibitem{perona1998factorization}
Perona, P., Freeman, W.: A factorization approach to grouping, \emph{LNCS},
  vol. 1406, pp. 655--670.
\newblock Springer Berlin Heidelberg (1998)

\bibitem{petitot1999vers}
Petitot, J., Tondut, Y.: Vers une neurog{\'e}om{\'e}trie. fibrations
  corticales, structures de contact et contours subjectifs modaux.
\newblock Math. Inform. Sci. Humaines \textbf{145}, 5--101 (1999)

\bibitem{poon2007live}
Poon, K., Hamarneh, G., Abugharbieh, R.: Live-vessel: Extending livewire for
  simultaneous extraction of optimal medial and boundary paths in vascular
  images, \emph{LNCS}, vol. 4792, pp. 444--451.
\newblock Springer Berlin Heidelberg (2007)

\bibitem{quek2001vessel}
Quek, F.K., Kirbas, C.: Vessel extraction in medical images by wave-propagation
  and traceback.
\newblock {IEEE} T. Med. Imaging \textbf{20}(2), 117--131 (2001)

\bibitem{robert2013monte}
Robert, C., Casella, G.: Monte Carlo statistical methods.
\newblock Springer Science \& Business Media (2013)

\bibitem{sanguinetti2008image}
Sanguinetti, G., Citti, G., Sarti, A.: Image completion using a diffusion
  driven mean curvature flow in a sub-{R}iemannian space.
\newblock In: Proceedings of the Third Int. Conf. on Computer Vision Theory and
  Applications (VISIGRAPP 2008), pp. 46--53 (2008)

\bibitem{sanguinetti2010model}
Sanguinetti, G., Citti, G., Sarti, A.: A model of natural image edge
  co-occurrence in the rototranslation group.
\newblock J. Vision \textbf{10}(14), 37 (2010)

\bibitem{sarti2015constitution}
Sarti, A., Citti, G.: The constitution of visual perceptual units in the
  functional architecture of {V1}.
\newblock J. Comput. Neurosci. \textbf{38}(2), 285--300 (2015)

\bibitem{sarti2008symplectic}
Sarti, A., Citti, G., Petitot, J.: The symplectic structure of the primary
  visual cortex.
\newblock Biol. Cybern. \textbf{98}(1), 33--48 (2008)

\bibitem{shi2000normalized}
Shi, J., Malik, J.: Normalized cuts and image segmentation.
\newblock {IEEE} T. Pattern. Anal. \textbf{22}(8), 888--905 (2000)

\bibitem{smith2004retinal}
Smith, W., Wang, J.J., Wong, T.Y., Rochtchina, E., Klein, R., Leeder, S.R.,
  Mitchell, P.: Retinal arteriolar narrowing is associated with 5-year incident
  severe hypertension. {T}he {B}lue {M}ountains {E}ye {S}tudy.
\newblock Hypertension \textbf{44}(4), 442--447 (2004)

\bibitem{staal2004}
Staal, J., Abr{\`a}moff, M.D., Niemeijer, M., Viergever, M.A., van Ginneken,
  B.: Ridge-based vessel segmentation in color images of the retina.
\newblock {IEEE} T. Med. Imaging \textbf{23}(4), 501--509 (2004)

\bibitem{tabish2007diabetes}
Tabish, S.A.: Is diabetes becoming the biggest epidemic of the twenty-first
  century?
\newblock Int. J. Health Sci. \textbf{1}(2), V (2007)

\bibitem{viswanath2003diabetic}
Viswanath, K., McGavin, D.M.: Diabetic retinopathy: clinical findings and
  management.
\newblock Community Eye Health \textbf{16}(46), 21 (2003)

\bibitem{von2007tutorial}
Von~Luxburg, U.: A tutorial on spectral clustering.
\newblock Stat. Comput. \textbf{17}(4), 395--416 (2007)

\bibitem{wagemans2012century}
Wagemans, J., Elder, J.H., Kubovy, M., Palmer, S.E., Peterson, M.A., Singh, M.,
  von~der Heydt, R.: A century of {G}estalt psychology in visual perception: I.
  perceptual grouping and figure--ground organization.
\newblock Psychol. Bull. \textbf{138}(6), 1172 (2012)

\bibitem{wasan1995vascular}
Wasan, B., Cerutti, A., Ford, S., Marsh, R.: Vascular network changes in the
  retina with age and hypertension.
\newblock J. Hypertens \textbf{13}(12), 1724--1728 (1995)

\bibitem{weiss1999segmentation}
Weiss, Y.: Segmentation using eigenvectors: a unifying view.
\newblock In: The proceedings of the Seventh IEEE International Conference on
  Computer vision, vol.~2, pp. 975--982 (1999)

\bibitem{wertheimer1938laws}
Wertheimer, M.: Laws of organization in perceptual forms.
\newblock In: W.~Ellis (ed.) A {S}ource {B}ook of {G}estalt {P}sychology, pp.
  71--88. Routledge and Kegan Paul (1938)

\bibitem{williams1997stochastic}
Williams, L.R., Jacobs, D.W.: Stochastic completion fields: A neural model of
  illusory contour shape and salience.
\newblock Neural. Comput. \textbf{9}(4), 837--858 (1997)

\bibitem{wong2007eye}
Wong, T., Mitchell, P.: The eye in hypertension.
\newblock Lancet \textbf{369}(9559), 425--435 (2007)

\bibitem{wong2001retinal}
Wong, T.Y., Klein, R., Couper, D.J., Cooper, L.S., Shahar, E., Hubbard, L.D.,
  Wofford, M.R., Sharrett, A.R.: Retinal microvascular abnormalities and
  incident stroke: the atherosclerosis risk in communities study.
\newblock Lancet \textbf{358}(9288), 1134--1140 (2001)

\bibitem{xu2010novel}
Xu, L., Luo, S.: A novel method for blood vessel detection from retinal images.
\newblock Biomed. Eng. Online \textbf{9}(1), 14 (2010)

\bibitem{xu2011vessel}
Xu, X., Niemeijer, M., Song, Q., Sonka, M., Garvin, M.K., Reinhardt, J.M.,
  Abr{\`a}moff, M.D.: Vessel boundary delineation on fundus images using
  graph-based approach.
\newblock {IEEE} T. Med. Imaging \textbf{30}(6), 1184--1191 (2011)

\bibitem{zhang2014numerical}
Zhang, J., Duits, R., Sanguinetti, G., ter Haar~Romeny, B.M.: Numerical
  approaches for linear left-invariant diffusions on {SE(2)}, their comparison
  to exact solutions, and their applications in retinal imaging.
\newblock arXiv preprint:1403.3320  (2014)

\bibitem{zucker2006differential}
Zucker, S.: Differential geometry from the {F}renet point of view: boundary
  detection, stereo, texture and color.
\newblock In: Handbook of Mathematical Models in Computer Vision, pp. 357--373.
  Springer (2006)

\end{thebibliography}

\end{document}